%% file: full.tex
\documentclass[10pt,twocolumn,letterpaper]{article}

\usepackage[pagenumbers]{cvpr} 

\input{preamble}

\definecolor{cvprblue}{rgb}{0.21,0.49,0.74}
\usepackage[pagebackref,breaklinks,colorlinks,allcolors=cvprblue]{hyperref}

\def\confName{CVPR}
\def\confYear{2027}

\title{ActMem: From Latent Memory Reasoning to Grounded Robotic Action}

\author{Linfeng Dong$^{1}$\quad Jianwei Niu$^{1}$\quad Jiuyi Zhao$^{1}$\quad Ruibing Feng$^{1}$\quad Xinghao Wu$^{1}$\\
Wu Yang$^{2}$\quad Shuai Tao$^{2}$\quad Wulong Liu$^{2}$\quad Mu Li$^{1}$\\[4pt]
$^{1}$Beihang University\qquad $^{2}$Beta Infinity
}

\begin{document}
\maketitle
\suppressfloats[t]
\input{sec/0_abstract_arxiv}
\input{fig/teaser}
\input{sec/1_intro}
\input{sec/2_related}
\input{sec/3_method}
\input{sec/4_experiments}
\input{sec/5_conclusion}
\input{sec/llm_statement}
{
    \small
    \bibliographystyle{ieeenat_fullname}
    \bibliography{main}
}

\clearpage
\appendix
\input{sec/X_suppl}

\end{document}

%% file: preamble.tex
\usepackage{cuted} 

\usepackage{currfile} 

\usepackage{caption} 
\usepackage[protrusion=true,expansion=true]{microtype}

\usepackage{booktabs}
\usepackage{multirow}
\usepackage{colortbl}
\usepackage{graphicx}
\usepackage{pgfplots}
\pgfplotsset{compat=1.18}
\usepgfplotslibrary{groupplots}
\usetikzlibrary{patterns,positioning}
\definecolor{evkGreen}{HTML}{A9D18E}
\definecolor{evkBlue}{HTML}{9DC3E6}
\definecolor{evkGold}{HTML}{FFD966}
\definecolor{evkGray}{HTML}{BFBFBF}

%% file: sec/0_abstract_arxiv.tex
\begin{abstract}
Vision-Language-Action (VLA) models have shown promising results in real-world long-horizon manipulation, where memory plays a crucial role.
However, existing memory-augmented VLAs either retrieve past observations without reasoning about what they imply, or reason over the full history in a shared context where it crowds out current perception.
Cognitive science treats them as distinct systems: memory reconstructs past experience to decide what to do, while perception guides how to act.
We therefore propose \textbf{ActMem}, an in-context memory framework that decouples memory reasoning from perceptual grounding within a single vision-language backbone.
Built on a vision-language model with native video understanding, ActMem infers the current subtask from a timestamped interaction history and keeps the hidden states formed during this inference as a \textbf{Latent Memory Chain-of-Thought (Memory CoT)}, which conditions a flow-matching action expert.
\textbf{Perception Attention} further lets the expert attend directly to the current view.
To keep inference fast under long histories, \textbf{Efficient Memory CoT Inference} reuses computation across consecutive decisions without changing the output.
ActMem achieves \textbf{71.1\%} task success and \textbf{80.8\%} completion success on RoboMemArena, surpassing the strongest baseline by \textbf{7.5} and \textbf{8.7} percentage points, and \textbf{73.8\%} average success on MIKASA-Robo, the best among all compared methods.
In the real world, it reaches \textbf{68.8\%} average success on exogenous, endogenous, and traceless memory tasks, \textbf{10.0} percentage points above the strongest memory-augmented VLA.
Code and checkpoints will be released at \url{https://github.com/LigphiDonk/ActMem} and \url{https://huggingface.co/donk111/ActMem}.
\end{abstract}

%% file: fig/teaser.tex
\begin{figure}[t]
\centering
\includegraphics[width=0.9\linewidth]{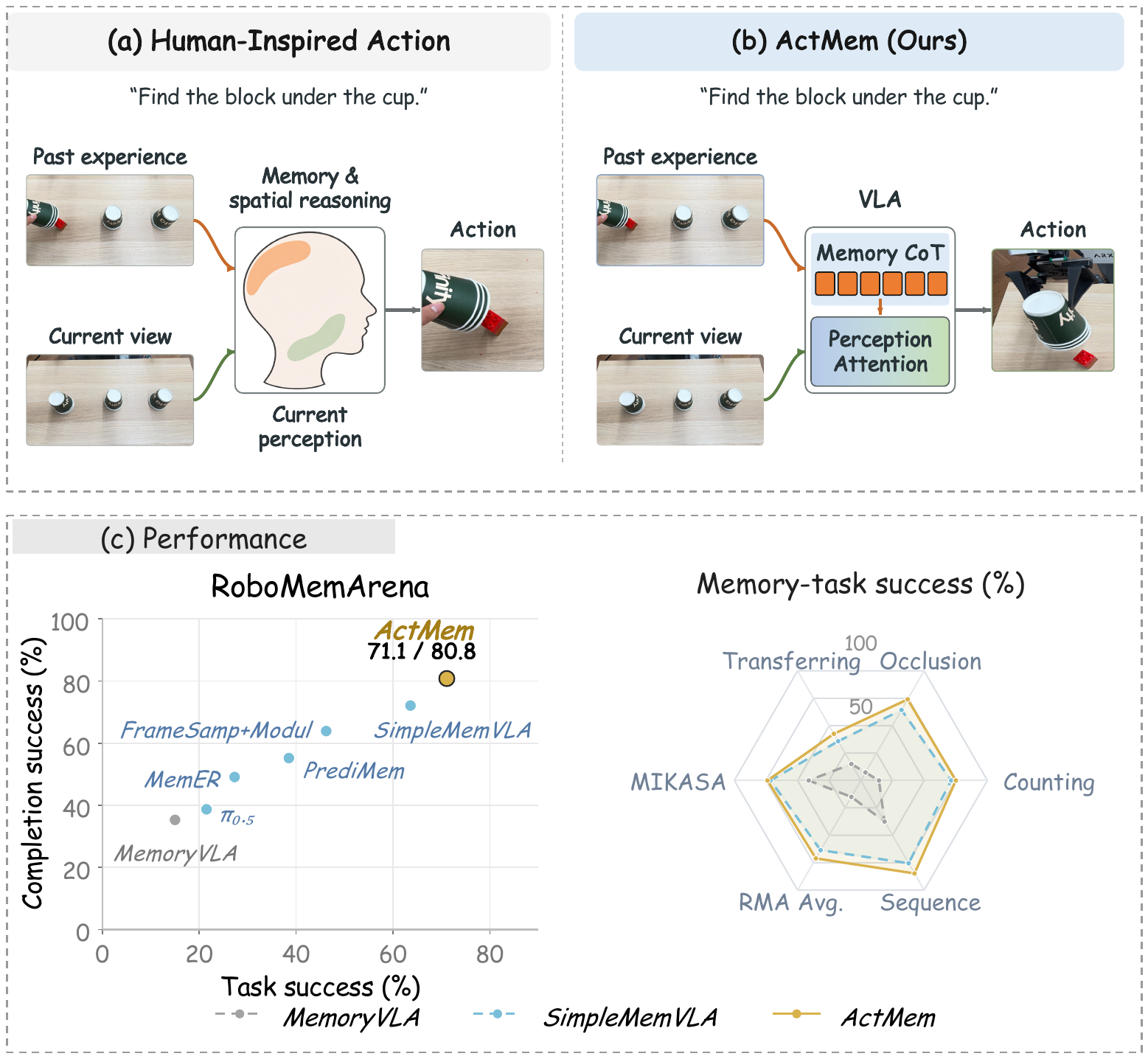}
\caption{\textbf{ActMem: history becomes action.} (a) Human-inspired action: past experience is reconstructed into memory and spatial reasoning that determines what to do, while current perception determines how to act. (b) ActMem realizes this division inside a VLA: latent memory reasoning in the vision-language backbone produces Memory CoT states that condition the action expert, and Perception Attention grounds the action in the current view. (c) RoboMemArena completion versus task success (left) and memory-task success across RoboMemArena categories and MIKASA-Robo (right).}
\label{fig:teaser}
\end{figure}

%% file: sec/1_intro.tex
\section{Introduction}
\label{sec:intro}

Vision-Language-Action (VLA) models have become a leading approach to generalist robot manipulation~\cite{rt2,openvla,pi0,gr00t} and have begun to tackle long-horizon tasks in the real world~\cite{pi05,hirobot}.
Most of them, however, process each frame in isolation~\cite{openvla,pi0,pi05}, treating the current observation as the full state of the task.
This assumption breaks once the scene no longer bears a trace of what has happened: after a block is hidden under a cup, that cup looks like any other, yet the correct action has changed.
Which subgoals are complete, or where an occluded object lies, cannot be read from the current frame~\cite{kaelbling1998pomdp,mikasa}.

Recent memory-augmented VLAs therefore incorporate history, but they fall short in one of two ways.
Memory-bank and retrieval policies such as MemoryVLA~\cite{memoryvla}, MEM~\cite{mem}, and MemER~\cite{memer} keep the current view intact but store past observations, leaving the action module to work out what they imply.
Native-video policies such as SimpleMemVLA~\cite{simplememvla} do reason about history-dependent subtasks, but in the same context that must also hold the current view, so the visual detail needed to act competes with an ever-growing history.
The challenge is thus to reason over the full history without crowding out current perception.

Our key observation is that a memory-dependent decision needs not a past observation itself, but what the history implies for the current step.
Past frames record how the scene looked, but decision-relevant information such as task progress and operation counts appears in no single frame, only in the relations among frames~\cite{mikasa,robomemarena,robomme,rmbench}.
When a robot pulls tissues from a box, no image shows how many have been pulled: the count is never observed, only inferred, so memory must reason over history rather than store it.

Such inferences and the current view are different kinds of information.
Cognitive science has long assigned them to distinct systems: memory reconstructs past experience to determine what to do~\cite{schacter2007}, while perception guides how to act~\cite{goodale1992}.
This motivates a simple principle: a VLA should keep what it has experienced separate from what it perceives.
Experience is what must be inferred from interaction history, such as progress, repetition counts, and where a now-hidden object was placed; perception is what the current view shows directly, such as object pose, geometry, and the gripper's relation to the target, which determine the next motion.
The boundary is not between past and present frames, but between what must be inferred and what can be observed.

Following this principle, we present \textbf{ActMem}, an in-context memory framework that reasons over experience in latent space and grounds action in current perception (Fig.~\ref{fig:teaser}).
A pretrained vision-language model reads the timestamped interaction history, the current observation, and the instruction in one context, and autoregressively infers the current subtask.
We keep the contextualized hidden states of these subtask tokens as \textbf{Latent Memory Chain-of-Thought (Memory CoT)} and condition a flow-matching action expert on them.
Because these states are formed after the backbone has processed the full history, they carry inferences that the subtask text alone does not preserve.
\textbf{Perception Attention} additionally lets the expert attend directly to contextualized visual features of the current view.
Memory CoT thus carries the reasoning over history, and Perception Attention gives the expert a direct path to the current view, so that perception no longer competes with history.

We evaluate ActMem on two memory-dependent simulation benchmarks, RoboMemArena and MIKASA-Robo, and on four real-world tasks covering exogenous, endogenous, and traceless memory.
On RoboMemArena, ActMem reaches 71.1\% task success and 80.8\% completion success, surpassing the strongest baseline by 7.5 and 8.7 percentage points.
It also achieves the best average success on MIKASA-Robo (73.8\%) and 68.8\% average success in the real world, 10.0 percentage points above the strongest memory-augmented VLA.
By reusing computation across consecutive decisions without changing the output, ActMem keeps per-decision latency below 0.6\,s even with long histories.

Our main contributions are summarized as follows:
\begin{itemize}
    \item \textbf{A memory principle for VLAs.} We identify that existing memory-augmented VLAs either retrieve history without reasoning about it or reason over it at the cost of current perception, and propose, consistent with the distinction between memory and perception in cognitive science, that a VLA should separate what it has experienced from what it perceives: memory reasons over history about what to do, while perception guides how to act in the current view.
    \item \textbf{A memory-reasoning VLA architecture.} ActMem conditions a flow-matching action expert on \textbf{Latent Memory CoT} from history-conditioned subtask inference, grounds generation in the current view via \textbf{Perception Attention}, and uses \textbf{Efficient Memory CoT Inference} to stay fast under long histories.
    \item \textbf{Strong performance across memory types.} ActMem achieves the best average success on RoboMemArena and MIKASA-Robo, and outperforms prior memory-augmented VLAs on four real-world tasks spanning exogenous, endogenous, and traceless memory.
\end{itemize}

%% file: sec/2_related.tex
\section{Related Work}
\label{sec:related}

\noindent\textbf{Vision-Language-Action Models.}
Pretrained vision-language models (VLMs)~\cite{paligemma,qwen25vl,qwen3vl} provide the semantic and visual knowledge on which VLA models build to map visual observations and language instructions to robot actions~\cite{rt2,openvla,octo,pi0,cogact,spatialvla,gr00t,openvlaoft,starvla}.
Early approaches formulate control as autoregressive prediction over discretized action tokens~\cite{rt2,openvla}, while more recent methods employ diffusion or flow-matching action experts to generate continuous motor commands~\cite{pi0,cogact,gr00t}.
However, most VLAs condition on the current observation alone, even when their VLM backbones can process video, and thus cannot recover task-relevant events that are no longer visible.

\vspace{0.45em}
\noindent\textbf{Memory for Robotic Manipulation.}
Recent benchmarks formalize memory-dependent manipulation, where the correct action depends on events no longer visible in the current scene~\cite{mikasa,robomemarena,robomme,rmbench}.
Existing memory-augmented policies differ mainly in how they supply history: stacking past frames or motion features as temporal context~\cite{contextvla,hamlet,cronusvla,hifvla}, maintaining memory banks of past observations~\cite{memoryvla,memoryvlapp,mem,nativemem,sam2act}, retrieving task-relevant keyframes~\cite{memer,kemo,eventvla}, or encoding timestamped histories as native video~\cite{simplememvla}.
Memory has also been explored in world action models~\cite{memorywam,dimwam} and in-context or test-time adaptation~\cite{gmp,t2mem}.
In most of these approaches, history reaches the action module as observations, whether raw, compressed, or retrieved, so the module must still infer what that history implies for the current step.

\vspace{0.45em}
\noindent\textbf{Reasoning for Action Generation.}
Intermediate reasoning has been introduced to bridge high-level understanding and low-level control.
Hierarchical VLAs predict language subtasks that serve as instructions for the action policy~\cite{pi05,hirobot}, while action-space reasoning methods generate coarse reference trajectories and latent action priors~\cite{acotvla}.
These methods reason about the current task rather than over the interaction history, and language-based variants pass only the decoded text to the action generator.

More recently, native-video approaches infer history-dependent subtasks from timestamped interaction histories to guide a flow-matching action expert~\cite{simplememvla}, yet memory reasoning and current perception still reach the action expert through a shared context.
ActMem instead separates the roles of experience and perception: Latent Memory CoT carries the history-conditioned reasoning states to the action expert, while Perception Attention grounds action generation in the current view.

%% file: sec/3_method.tex
\section{Method}
\label{sec:method}

\input{fig/overview}

We present ActMem, an in-context memory framework that separates experience from perception, as illustrated in Fig.~\ref{fig:overview}.
A pretrained vision-language backbone infers the current subtask from the timestamped history; the resulting hidden states are kept as \textbf{Latent Memory CoT} to condition a flow-matching action expert, while \textbf{Perception Attention} grounds generation in the current view.

\subsection{Preliminaries and Architecture Overview}
\label{sec:preliminaries}

\noindent\textbf{Problem Formulation.}
We consider language-conditioned robotic manipulation under partial observability, where successful action generation may require information from previous interactions.
At each decision step $t$, the robot receives a language instruction $g$, a current visual observation $o_t$, and a proprioceptive state $s_t$.
In memory-dependent scenarios, the current observation alone may be insufficient to determine the appropriate action, as task-relevant events may have occurred earlier and are no longer directly observable.
We therefore denote the sampled interaction history as
\begin{equation}
\mathcal{H}_t=\{(o_{\tau_i},\tau_i)\}_{i=1}^{N_t},
\label{eq:history}
\end{equation}
where $o_{\tau_i}$ represents a historical visual observation and $\tau_i$ is its corresponding timestamp.
Given the interaction history and current observation, our objective is to learn a history-conditioned policy that generates a continuous action chunk:
\begin{equation}
A_t \sim \pi_\theta(\cdot \mid \mathcal{H}_t,o_t,s_t,g),
\label{eq:policy}
\end{equation}
where $A_t=(a_{t+1},\ldots,a_{t+H})$ denotes a sequence of robot actions over a prediction horizon $H$.

\noindent\textbf{Architecture Overview.}
As illustrated in Fig.~\ref{fig:overview}, ActMem consists of a pretrained vision-language backbone $f_\phi$ and a flow-matching action expert $f_\psi$.
The backbone processes the interaction history, current observations, and language instruction to construct a shared multimodal context.
From this context, ActMem derives two forms of action conditioning with distinct functional roles.
First, Latent Memory CoT preserves the contextualized representations formed during history-conditioned subtask generation, encoding experience such as task intent and progress as compact guidance for action selection.
Second, Perception Attention exposes contextualized visual features of the current view to intermediate action representations, supplying the perceptual information needed for motor execution.
The action expert integrates these signals with the current robot state to generate continuous action chunks.

\subsection{Latent Memory Chain-of-Thought}
\label{sec:subtask}
\label{sec:nativevideo}

Rather than compressing or selectively retaining past observations, ActMem lets a pretrained vision-language model with native video understanding reason directly over the interaction history, and keeps the states formed during this reasoning as the memory condition for action generation.

\noindent\textbf{Native-Video Context.}
Given the interaction history $\mathcal{H}_t$, we construct a timestamped front-camera video $V_t=[(\tau_1,v_{\tau_1}),\ldots,(\tau_{N_t},v_{\tau_{N_t}})]$, so the backbone can reason about the temporal order of events and the progression of the task.
Together with the current wrist-camera view $w_t$, which serves as $o_t$ in Eq.~\eqref{eq:policy}, and the instruction $g$, it forms a single multimodal input $X_t=\operatorname{Tokenize}(V_t,w_t,g)$.
We employ Qwen3-VL as the backbone $f_\phi$, which encodes positions across visual and textual tokens with multimodal rotary position embeddings (M-RoPE), yielding contextualized representations $Z_t=f_\phi(X_t)\in\mathbb{R}^{L_t\times d}$.

\noindent\textbf{History-Conditioned Subtask Reasoning.}
From $X_t$, the backbone autoregressively infers the subtask $\ell_t=(\ell_{t,1},\ldots,\ell_{t,M_t})$ for the current stage of the task:
\begin{equation}
p_\phi(\ell_t\mid X_t)=\prod_{j=1}^{M_t}p_\phi(\ell_{t,j}\mid X_t,\ell_{t,<j}).
\label{eq:subtask}
\end{equation}
Because this inference is conditioned on the history, it can distinguish visually similar situations that require different actions, such as which drawers have already been inspected.

\noindent\textbf{Memory CoT Conditioning.}
Rather than conditioning the action expert on a separate encoding of the subtask text or on the full history tokens, we use the final-layer hidden states of the generated subtask span as \textbf{Latent Memory CoT}.
Letting $\mathcal{I}_t$ denote the positions of $\ell_t$, we define
\begin{equation}
C_t=\left\{(h_i,e_i,p_i)\mid i\in\mathcal{I}_t\right\},
\label{eq:memcot}
\end{equation}
where $h_i$, $e_i$, and $p_i$ are the hidden state, token embedding, and M-RoPE position of the $i$-th subtask token.
As shown in Fig.~\ref{fig:overview}(b), the subtask tokens follow the history, the current view, and the instruction, so under the causal mask their hidden states summarize what the backbone has inferred from the full context.
The action expert receives only $C_t$, whose length depends on the subtask rather than on the number of historical frames.

\subsection{Perception Attention for Action Generation}
\label{sec:present}

While history-conditioned subtask representations abstract task intent and progress, continuous action generation also depends on the perceptual details of the scene.
We therefore introduce \textbf{Perception Attention}, which lets intermediate action representations attend to contextualized visual features during flow-matching generation, grounding motor execution in richer perceptual context.

\noindent\textbf{Visual Conditioning.}
Given the multimodal representations $Z_t$ produced by the vision-language backbone, let $\mathcal{J}_t$ denote the positions of the current-view tokens $w_t$ in $X_t$.
As for $C_t$, we take the final-layer hidden states at these positions as the visual features:
\begin{equation}
P_t=\{h_i \mid i\in\mathcal{J}_t\}\in\mathbb{R}^{N_p\times d},
\label{eq:percep}
\end{equation}
where $N_p=|\mathcal{J}_t|$ is the number of current-view tokens.
Since these tokens are computed after the backbone has processed the interaction history, they remain history-contextualized.
Unlike the compact Memory CoT conditioning sequence $C_t$, which provides task-level guidance, $P_t$ preserves a token-level interface to the current scene.
These representations allow the action expert to incorporate perceptual information alongside history-conditioned subtask representations.

\noindent\textbf{Gated Visual Cross-Attention.}
We implement Perception Attention as gated cross-attention at selected layers of the flow-matching Diffusion Transformer (DiT).
At the $l$-th selected layer, the intermediate action representations $U_t^{(l)}$ query the visual features $P_t$:
\begin{equation}
R_t^{(l)}
=
\operatorname{CrossAttn}
\left(
Q=U_t^{(l)},\,
K=P_t,\,
V=P_t
\right).
\label{eq:crossattn}
\end{equation}
The resulting visual information is incorporated into the action representations through a learnable residual gate:
\begin{equation}
\widetilde{U}_t^{(l)}
=
U_t^{(l)}
+
\tanh(\alpha_l)R_t^{(l)},
\label{eq:gate}
\end{equation}
where $\alpha_l$ is initialized to zero.
This initialization preserves the original action representations at the beginning of training, while allowing the contribution of visual information to be learned progressively.
In our implementation, Perception Attention is inserted at layers $\{3,7,11,15\}$ of a 16-layer DiT, while retaining the original subtask-conditioning mechanism.

\noindent\textbf{Flow-Matching Action Generation.}
The action expert predicts a velocity field $v_\psi(A_t^\lambda,\lambda\mid C_t,P_t,s_t)$ over noisy action chunks $A_t^\lambda$ at flow timestep $\lambda$, where $C_t$ serves as the conditioning sequence and $P_t$ enters through Perception Attention.
Integrating this field from Gaussian noise yields the action chunk $A_t$.

\input{fig/realworld}

\subsection{Policy Learning and Inference}
\label{sec:learning}

\noindent\textbf{Joint Policy Learning.}
We train ActMem on manipulation demonstrations containing visual observations, language instructions, subtask annotations, and continuous robot actions.
The vision-language backbone and flow-matching action expert are jointly optimized to associate historical context with the appropriate subtask while learning to generate the corresponding robot actions.
The overall training objective combines a subtask prediction loss and an action generation loss:
\begin{equation}
\mathcal{L}
=
\lambda_{\mathrm{sub}}\mathcal{L}_{\mathrm{sub}}
+
\lambda_{\mathrm{act}}\mathcal{L}_{\mathrm{act}},
\label{eq:loss}
\end{equation}
where $\lambda_{\mathrm{sub}}$ and $\lambda_{\mathrm{act}}$ control the relative contributions of the two objectives.
For history-conditioned subtask prediction, we employ an autoregressive cross-entropy loss over the annotated subtask sequence:
\begin{equation}
\mathcal{L}_{\mathrm{sub}}
=
-\frac{1}{M_t}
\sum_{j=1}^{M_t}
\log p_\phi
\left(
\ell^*_{t,j}
\mid X_t,\ell^*_{t,<j}
\right),
\label{eq:subloss}
\end{equation}
where $\ell_t^*$ denotes the ground-truth subtask description.
This supervision encourages the vision-language backbone to infer the appropriate task stage from the multimodal context, providing history-conditioned representations for subsequent action generation.
As illustrated in Fig.~\ref{fig:overview}(a), the contextualized states associated with the subtask tokens form the Memory CoT condition $C_t$.

\noindent\textbf{Flow-Matching Objective.}
We train the action expert with conditional flow matching~\cite{pi0}.
Given a demonstrated chunk $A_t$, noise $\epsilon\sim\mathcal{N}(0,I)$, and $\lambda\sim\operatorname{Beta}(1.5,1.0)$, we form $A_t^\lambda=(1-\lambda)A_t+\lambda\epsilon$ and regress the target velocity $\epsilon-A_t$:
\begin{equation}
\mathcal{L}_{\mathrm{act}}=\mathbb{E}\,\big\|v_\psi(A_t^\lambda,\lambda\mid C_t,P_t,s_t)-(\epsilon-A_t)\big\|_2^2.
\label{eq:fmloss}
\end{equation}

\noindent\textbf{Closed-Loop Inference.}
During deployment, ActMem maintains a history buffer that accumulates visual observations throughout task execution.
At each replanning step, the sampled historical frames, current wrist-camera observation, and language instruction are provided to the vision-language backbone to predict the current subtask.
From the resulting context we extract the Memory CoT representations $C_t$ and the visual features $P_t$, and the action expert generates an action chunk conditioned on them and the current proprioceptive state $s_t$.
In our implementation, the policy predicts 16 actions using 10 flow-matching integration steps and executes the first 10 actions before acquiring new observations and replanning.
This receding-horizon procedure enables ActMem to continually update its history-conditioned task intent and motor actions as the interaction unfolds.

\noindent\textbf{Efficient Memory CoT Inference.}
Consecutive decisions differ only slightly: the history gains at most one new frame chunk, and the subtask typically persists over several decisions (Fig.~\ref{fig:overview}(c)).
Standard inference nonetheless re-encodes the full history, decodes the subtask token by token, and runs a second backbone pass to obtain $C_t$, so its cost grows with the history length.
We remove this redundancy through exact reuse, producing the same result as standard inference up to numerical precision.
Since the vision encoder attends only within each two-frame chunk and the history precedes the current view and instruction while only being appended, we cache visual features per chunk and reuse the key-value (KV) cache of the unchanged history prefix, so each decision encodes only the newly added chunk and prefills only the remaining tokens.
For the subtask, we take the previous subtask $\hat{\ell}=\ell_{t-1}$ as a draft, append it to the prompt, and obtain the greedy predictions at all draft positions in a single parallel pass:
\begin{equation}
\tilde{\ell}_j=\arg\max_{v}\, p_\phi\!\left(v\mid X_t,\hat{\ell}_{<j}\right),\qquad j=1,\ldots,|\hat{\ell}|.
\label{eq:draft}
\end{equation}
The draft is accepted up to its first mismatch,
\begin{equation}
k=\max\left\{j\ge 0 \mid \tilde{\ell}_{1:j}=\hat{\ell}_{1:j}\right\}.
\label{eq:accept}
\end{equation}
If $k=|\hat{\ell}|$, the previous subtask is kept without any decoding step; otherwise, the KV cache is truncated after $\hat{\ell}_{1:k}$ and greedy decoding resumes from $\tilde{\ell}_{k+1}$.
Following the verification rule of speculative decoding~\cite{leviathan2023speculative}, this yields exactly the greedy-decoded subtask.
Because the draft is re-processed under the updated context, the hidden states at its positions are recomputed in this pass and directly form $C_t$, which removes the second backbone pass and avoids stale latent conditions.
As a result, each decision encodes a single new chunk, and the previous subtask is accepted without decoding in most decisions (see supplementary material).

%% file: fig/overview.tex
\begin{figure*}[t]
\centering
\includegraphics[width=0.97\linewidth]{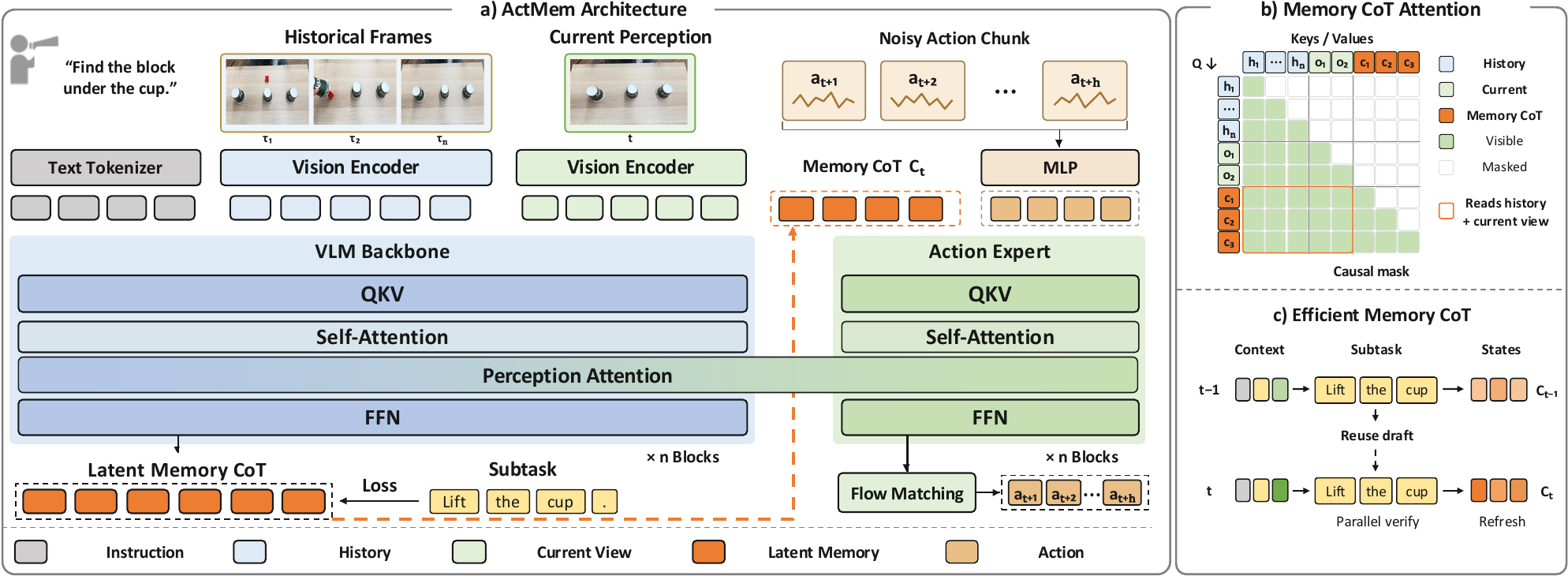}
\caption{Overview of ActMem. (a) A pretrained vision-language backbone (Qwen3-VL) encodes the instruction, the timestamped historical frames, and the current perception, and performs latent memory reasoning inside the backbone to produce the subtask together with its Latent Memory CoT states $C_t$, which condition a flow-matching action expert; Perception Attention additionally routes the current visual features to the action expert. (b) Inside the backbone, the subtask tokens that form Memory CoT follow the history and the current view in the sequence, so under the causal mask they read both, and their hidden states become $C_t$; the action expert receives only $C_t$, not the history tokens. (c) Efficient Memory CoT Inference. Consecutive decisions usually share the same subtask, so the previous subtask is reused as a draft and verified against the updated context in a single parallel pass. If it is accepted, its hidden states are recomputed to give the new $C_t$, with no token-by-token decoding.}
\label{fig:overview}
\end{figure*}

%% file: fig/realworld.tex
\begin{figure*}[t]
\centering
\vspace{-6pt}
\includegraphics[width=0.95\linewidth]{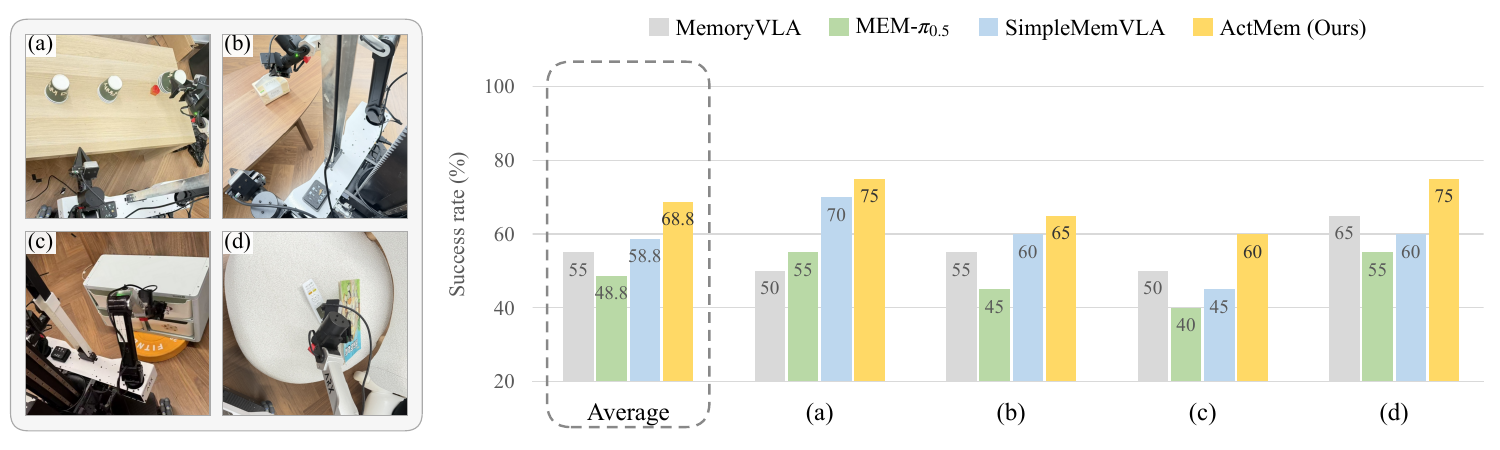}
\caption{Visualization of real-world tasks and comparison of real-world performance.
\textbf{Left}: (a) Cup-and-Block, (b) Tissue Counting, (c) Cabinet Search, and
(d) Scissor Retrieval, spanning exogenous, endogenous, and traceless memory.
\textbf{Right}: success rates of MemoryVLA, MEM-$\pi_{0.5}$, SimpleMemVLA, and ActMem.
Each task is evaluated over 40 trials, and success rates are computed accordingly.}
\label{fig:realworld}
\vspace{-8pt}
\end{figure*}

%% file: sec/4_experiments.tex
\begin{table*}[t]
\centering
\caption{Comparison on RoboMemArena. TSR and CSR (\%) over all 26 tasks under the official protocol (26 tasks $\times$ 51 trials). ActMem is ours; remaining baselines are taken from prior papers~\cite{simplememvla,robomemarena}. Best in each column in \textbf{bold}.}
\label{tab:robomemarena}
\small
\setlength{\tabcolsep}{6pt}
\begin{tabular}{lcccccccccc}
\toprule
\multirow{2}{*}{Method} & \multicolumn{2}{c}{Transferring} & \multicolumn{2}{c}{Occlusion} & \multicolumn{2}{c}{Counting} & \multicolumn{2}{c}{Sequence} & \multicolumn{2}{c}{Average} \\
\cmidrule(lr){2-3}\cmidrule(lr){4-5}\cmidrule(lr){6-7}\cmidrule(lr){8-9}\cmidrule(lr){10-11}
 & TSR & CSR & TSR & CSR & TSR & CSR & TSR & CSR & TSR & CSR \\
\midrule
\multicolumn{11}{c}{\textit{Reference}} \\
Qwen3-VL-8B (frozen)~\cite{robomemarena} & 15.0 & 34.6 & 0.0 & 6.8 & 9.3 & 44.6 & 7.5 & 39.2 & 6.0 & 26.2 \\
GPT-5.4~\cite{robomemarena} & 13.8 & 32.9 & 1.8 & 9.2 & 12.9 & 50.7 & 15.0 & 47.3 & 8.7 & 30.5 \\
\midrule
\multicolumn{11}{c}{\textit{Reactive / short-horizon VLAs}} \\
$\pi_{0.5}$~\cite{pi05} & 20.0 & 42.8 & 12.7 & 17.2 & 14.3 & 50.9 & 60.0 & 71.6 & 21.5 & 38.7 \\
HiF-VLA~\cite{hifvla} & 17.5 & 38.9 & 12.7 & 27.1 & 8.6 & 45.9 & 42.5 & 70.2 & 16.9 & 39.8 \\
\midrule
\multicolumn{11}{c}{\textit{Memory-augmented VLAs}} \\
MemoryVLA~\cite{memoryvla} & 15.0 & 37.2 & 7.3 & 13.1 & 14.3 & 55.1 & 37.5 & 65.2 & 15.0 & 35.3 \\
MemER~\cite{memer} & 20.0 & 36.1 & 16.4 & 33.2 & 27.1 & 65.1 & 65.0 & 79.1 & 27.3 & 49.1 \\
FrameSamp+Modul~\cite{robomme} & \textbf{63.8} & \textbf{72.1} & 39.1 & 56.5 & 31.4 & 57.8 & 73.8 & 86.9 & 46.2 & 63.9 \\
PrediMem (Qwen3-1.7B)~\cite{robomemarena} & 15.0 & 31.6 & 7.3 & 20.8 & 28.6 & 60.2 & 50.0 & 73.9 & 19.9 & 41.4 \\
PrediMem (Qwen3-4B)~\cite{robomemarena} & 20.0 & 42.1 & 18.2 & 34.7 & 38.6 & 64.9 & 65.0 & 84.6 & 31.9 & 51.7 \\
PrediMem~\cite{robomemarena} & 22.5 & 45.2 & 27.3 & 38.4 & 45.7 & 69.3 & 72.5 & 89.5 & 38.5 & 55.2 \\
SimpleMemVLA~\cite{simplememvla} & 35.8 & 37.4 & 64.3 & 76.8 & 71.4 & 79.3 & 75.5 & 81.1 & 63.6 & 72.1 \\
\midrule
\rowcolor{evkBlue!45}ActMem (Ours) & 42.6 & 46.3 & \textbf{74.0} & \textbf{88.3} & \textbf{75.1} & \textbf{82.2} & \textbf{84.8} & \textbf{92.4} & \textbf{71.1} & \textbf{80.8} \\
\bottomrule
\end{tabular}
\end{table*}

\section{Experiments}
\label{sec:experiments}

We evaluate ActMem on simulated memory benchmarks and real-world manipulation tasks to answer four questions:
(1) How does ActMem compare with existing VLA policies on memory-dependent manipulation benchmarks?
(2) Can it handle different forms of historical dependence that require knowing task progress beyond the current observation?
(3) Where do its gains come from: the long interaction history, Latent Memory CoT, or Perception Attention?
(4) Does conditioning on a long history remain practical at inference time?
To answer these questions, we evaluate ActMem on RoboMemArena and MIKASA-Robo, run real-world experiments covering exogenous, endogenous, and traceless memory, and conduct controlled ablations and diagnostic analyses on RoboMemArena, together with a latency study under increasing history length.

\subsection{Experimental Setup}
\label{sec:setup}

\noindent\textbf{Benchmarks.}
We evaluate ActMem on two memory-dependent simulation benchmarks and four real-world manipulation tasks.
RoboMemArena~\cite{robomemarena} comprises 26 tasks in four categories (transferring, occlusion, counting, and sequence execution), in all of which the correct action depends on events that are no longer visible in the current observation.
MIKASA-Robo~\cite{mikasa} evaluates history-dependent manipulation on five tasks: ShellGame, Intercept, and RememberColor-3/5/9 (RC-3/5/9).
Following each benchmark's protocol, we run 51 trials per task on RoboMemArena and 100 trials per task on MIKASA-Robo, and evaluate a single trained model per benchmark.
On RoboMemArena we report task success rate (TSR), the fraction of episodes in which every subtask stage is completed, and completion success rate (CSR), the average fraction of stages completed per episode; on MIKASA-Robo we report the average task success rate.
The four real-world tasks span exogenous, endogenous, and traceless memory and are described in Sec.~\ref{sec:exp_real}.

\noindent\textbf{Baselines.}
For simulation, we compare against representative vision-language-action models and memory-augmented policies, reporting numbers from the corresponding publications or our own reproductions.
For the real-world experiments, we compare against MemoryVLA~\cite{memoryvla}, MEM-$\pi_{0.5}$~\cite{mem}, and SimpleMemVLA~\cite{simplememvla}.

\noindent\textbf{Implementation Details.}
ActMem is built upon the pretrained starVLA~\cite{starvla} framework, comprising a Qwen3-VL backbone and a flow-matching action expert.
We jointly fine-tune the vision-language backbone and action expert, conditioning action generation on history-dependent subtask representations and contextualized visual features through Perception Attention, which is incorporated at layers 3, 7, 11, and 15 of the 16-layer action expert.
We follow the optimizer and learning-rate schedule of starVLA.
Historical observations are sampled at 1\,fps over 126\,s for RoboMemArena and 20\,fps over 3\,s for MIKASA-Robo.
Training is conducted on 8 NVIDIA RTX PRO 6000 GPUs with a global batch size of 64 for 60k iterations, and we evaluate the final checkpoint.
During inference, the policy predicts 16-step action chunks using 10 flow-matching integration steps and executes the first 10 actions before replanning.
Further model and training hyperparameters are provided in the supplementary material.

\subsection{Memory-Dependent Simulation}
\label{sec:exp_memory}

\input{fig/attnvis}

\begin{table}[t]
\centering
\caption{MIKASA-Robo success rates (\%, 100 trials). GMP is a per-task non-VLA memory policy. $^{\dagger}$ our reproduction of the official SimpleMemVLA checkpoint; other VLA baselines from~\cite{memoryvlapp,mikasa,simplememvla}. Best in each column in \textbf{bold}.}
\label{tab:mikasa}
\resizebox{\linewidth}{!}{%
\begin{tabular}{lcccccc}
\toprule
Method & ShellGame & Intercept & RC-3 & RC-5 & RC-9 & Avg \\
\midrule
\multicolumn{7}{c}{\textit{Reference}} \\
GMP~\cite{gmp} & \textbf{98} & 83 & \textbf{80} & \textbf{61} & 17 & 67.8 \\
\midrule
\multicolumn{7}{c}{\textit{Vision-language-action models}} \\
CronusVLA~\cite{cronusvla} & 32 & 5 & 31 & 13 & 9 & 18.0 \\
SpatialVLA~\cite{spatialvla} & 23 & 27 & 27 & 17 & 11 & 21.0 \\
OpenVLA-OFT~\cite{openvlaoft} & 47 & 14 & 59 & 16 & 6 & 28.4 \\
$\pi_0$~\cite{pi0} & 33 & 42 & 35 & 22 & 15 & 29.4 \\
Octo~\cite{octo} & 46 & 39 & 45 & 17 & 11 & 31.6 \\
MemoryVLA~\cite{memoryvla} & 88 & 24 & 44 & 30 & 20 & 41.2 \\
MemoryVLA++~\cite{memoryvlapp} & 97 & 40 & 50 & 19 & 16 & 44.4 \\
SimpleMemVLA$^{\dagger}$~\cite{simplememvla} & 97 & 76 & 69 & 53 & 62 & 71.4 \\
\midrule
\rowcolor{evkBlue!45}ActMem (Ours) & 95 & \textbf{85} & 73 & 49 & \textbf{67} & \textbf{73.8} \\
\bottomrule
\end{tabular}}
\end{table}

\noindent\textbf{RoboMemArena.}
As shown in Table~\ref{tab:robomemarena}, ActMem achieves 71.1\% task success rate (TSR) and 80.8\% completion success rate (CSR), outperforming SimpleMemVLA by 7.5 and 8.7 percentage points, respectively.
Gains are largest in Occlusion and Sequence (+9.7 and +9.3 TSR over the best baseline), which depend most on earlier interactions.
ActMem trails FrameSamp+Modul on Transferring (42.6\% vs.\ 63.8\%), which relies more on fine-grained manipulation; we attribute this to visual detail competing with history, a weakness also seen in SimpleMemVLA, where Perception Attention already lifts Transferring from 35.8\% to 42.6\%.

\noindent\textbf{MIKASA-Robo.}
ActMem achieves an average success rate of 73.8\% on MIKASA-Robo (Table~\ref{tab:mikasa}), surpassing SimpleMemVLA by 2.4 percentage points.
Gains are largest on Intercept (85\% vs.\ 76\%) and RememberColor-9 (67\% vs.\ 62\%), where the action depends on a cue that is no longer visible, while ShellGame remains comparable to SimpleMemVLA (95\% vs.\ 97\%).
The same pattern appears on RoboMemArena: the advantage is largest when the current observation conceals the information needed to act.

\subsection{Real-Robot Manipulation}
\label{sec:exp_real}

We evaluate ActMem on four real-world manipulation tasks spanning three forms of memory dependence: \textbf{exogenous}, \textbf{endogenous}, and \textbf{traceless} memory (Fig.~\ref{fig:realworld}).
Exogenous-memory tasks require recalling events introduced by external agents: in \textit{Cup-and-Block}, the robot must identify the cup containing a previously hidden block, while in \textit{Scissor Retrieval}, it must locate scissors that are no longer visible.
Endogenous memory is evaluated through \textit{Cabinet Search}, where the robot must track previously inspected drawers to avoid redundant searches.
Finally, \textit{Tissue Counting} tests traceless memory, requiring the robot to count completed pulls despite nearly identical observations after each action.

\noindent\textbf{Real-World Results.}
Each task is evaluated over 40 trials.
As shown in Fig.~\ref{fig:realworld}, ActMem achieves an average success rate of 68.8\%, outperforming SimpleMemVLA (58.8\%) by 10.0 percentage points, MemoryVLA (55.0\%) by 13.8 points, and MEM-$\pi_{0.5}$ (48.8\%) by 20.0 points, and is the best method on all four tasks.
ActMem reaches 75\% on both Cup-and-Block and Scissor Retrieval, where the required information was introduced by an external agent and is no longer visible, and 65\% on Tissue Counting, where the observations after each pull are nearly identical.
Performance is lowest on Cabinet Search (60\%), which requires the policy to track the drawers it has already inspected.
These results demonstrate that ActMem handles different forms of historical dependence, while tracking self-generated task progress remains the most challenging.

\input{fig/pca_latency}

\subsection{Ablation Studies}
\label{sec:ablation}

We conduct ablation studies on RoboMemArena to examine whether each design of ActMem is necessary: the long interaction history, the Latent Memory CoT that conveys it, and Perception Attention.
All variants, including the full model, are trained for 40k iterations with the same data and schedule and evaluated on the 26 tasks with 10 trials per task, so absolute numbers are lower than in Table~\ref{tab:robomemarena}.

\begin{table}[t]
\centering
\caption{Ablation studies of ActMem on RoboMemArena (40k-iteration schedule, 26 tasks $\times$ 10 trials). We report task success rate (TSR) and completion success rate (CSR); higher values indicate better performance.}
\label{tab:ablation}
\small
\setlength{\tabcolsep}{4pt}
\begin{tabular}{lcc}
\toprule
Configuration & TSR (\%) $\uparrow$ & CSR (\%) $\uparrow$ \\
\midrule
w/o History & 17.7 & 34.6 \\
10\,s History & 31.2 & 47.7 \\
\midrule
Text-only Subtask & 40.4 & 56.9 \\
Full-token Conditioning & 38.8 & 54.6 \\
\midrule
w/o Perception Attention & 47.7 & 64.2 \\
\midrule
\rowcolor{evkBlue!45}ActMem (Full) & \textbf{52.3} & \textbf{68.5} \\
\bottomrule
\end{tabular}
\end{table}

\noindent\textbf{Interaction History.}
As shown in Table~\ref{tab:ablation}, removing historical observations reduces TSR from 52.3\% to 17.7\%, and keeping only the most recent 10\,s recovers just 31.2\%.
This indicates that the required information lies far back in the episode and that a short context cannot substitute for the full history.

\noindent\textbf{Latent Memory CoT.}
Given the same history, how it reaches the action expert matters.
Conditioning on the text embeddings of the generated subtask alone yields 40.4\% TSR, and letting the action expert read all history tokens directly yields 38.8\%, both below the 52.3\% of Latent Memory CoT.
Subtask text is therefore an incomplete memory signal: the action expert does not recover this progress from the raw history.

\noindent\textbf{Perception Attention.}
Removing Perception Attention reduces TSR from 52.3\% to 47.7\% and CSR from 68.5\% to 64.2\%.
This indicates that subtask-level guidance alone does not capture the visual information useful for continuous action generation.
The two conditions are complementary: Memory CoT selects what to do from experience, and Perception Attention supplies the current geometry with which to do it.

\subsection{Analysis}
\label{sec:analysis}

\noindent\textbf{Historical Evidence.}
Fig.~\ref{fig:attnvis} visualizes attention over historical frames in a representative RoboMemArena task.
At the decision step, ActMem attends most to the frame containing the task-relevant event, although this event is no longer visible in the current observation.

\noindent\textbf{What Latent Memory CoT Encodes.}
To test whether Memory CoT carries information beyond its text, we take the counting task \textit{Pour wine into the mug twice} and extract representations at the end of the first and the second pour for all 100 demonstrations (Fig.~\ref{fig:pca}).
The two moments are nearly indistinguishable in current visual features encoded before the history, whose class means have a cosine similarity of 0.99, whereas Memory CoT places them far apart (0.22; Fig.~\ref{fig:pca}).
The gap persists (0.25) when the same text ``pour the wine into the mug'' is forced for every sample, so the count is carried by the hidden states, not only by the words of the subtask.
In Fig.~\ref{fig:pca}, visual features of both moments collapse into a single cluster, whereas Memory CoT splits them into two groups along the count direction.
Thus, subtask inference does more than emit text: its hidden states keep a compact record of task progress that the action expert reads directly.

\subsection{Inference Efficiency}
\label{sec:exp_latency}

As shown in Fig.~\ref{fig:latency}, the per-decision latency of standard inference grows from 0.87\,s to 1.42\,s as the history lengthens from under 30 to over 90 frames, close to that of SimpleMemVLA (0.70\,s to 1.25\,s).
Efficient Memory CoT Inference (Sec.~\ref{sec:learning}) reduces it to 0.33--0.52\,s, $2.1\times$ to $2.8\times$ faster than standard inference and about $2.0\times$ to $2.5\times$ faster than SimpleMemVLA at every history length.
The gain comes from exactly reusing the history cache and the previous subtask (Sec.~\ref{sec:learning}), which leaves the output unchanged.
Since the history is only appended, the cost of each decision depends mainly on the new chunk rather than on the full history, so efficient latency rises by only 0.19\,s over the history range, versus 0.55\,s for standard inference.
$\pi_{0.5}$-family policies with a fixed-size memory run at 0.08--0.09\,s, so ActMem trades part of this speed for access to the full history.

%% file: fig/attnvis.tex
\begin{figure*}[t]
\centering
\includegraphics[width=\linewidth]{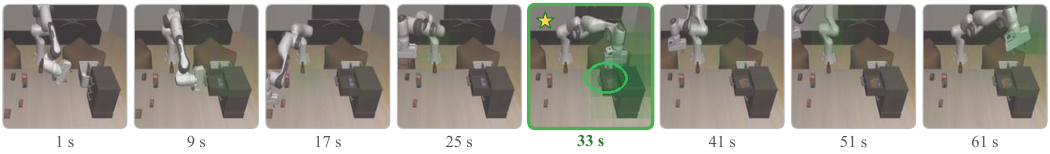}
\vspace{-14pt}
\caption{\textbf{Attention over historical frames.} History frames of the RoboMemArena task \textit{Put cookies into the top drawer and put chocolate into another drawer}, overlaid with ActMem's attention (green) at 61\,s (predicted subtask: ``close the middle drawer''). Attention is the causal softmax from the generated subtask tokens to the history tokens, averaged over subtask tokens, heads, and backbone layers 3, 7, \ldots, 31. The starred frame (33\,s) contains the task-relevant event, no longer visible at the decision step.}
\label{fig:attnvis}
\vspace{-6pt}
\end{figure*}

%% file: fig/pca_latency.tex

\begin{figure*}[t]
\centering
\includegraphics[width=\linewidth]{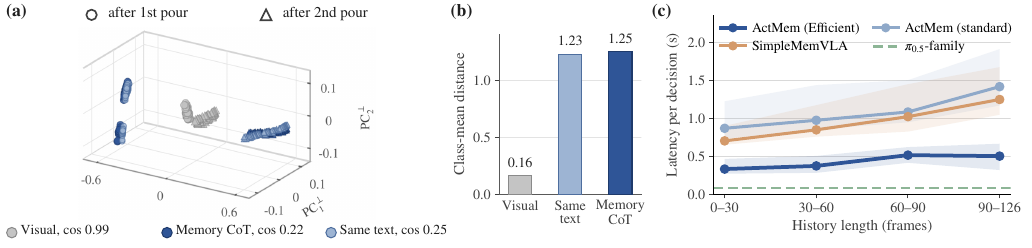}
\vspace{-8pt}
\caption{\textbf{(a) What Latent Memory CoT encodes.} Representations after the first (circles) and second (triangles) pour in \textit{Pour wine into the mug twice} (100 demonstrations); numbers give the cosine similarity between the two class means. Same text: ``pour the wine into the mug'' is teacher-forced for all samples. Residual axes are magnified for visibility.
\textbf{(b) Class-mean distance} between the two moments in each feature space, divided by the mean feature norm.
\textbf{(c) Inference latency} per decision (median, single GPU) versus history length on RoboMemArena; shaded: interquartile range; dashed: $\pi_{0.5}$-family models with fixed-size memory (see supplementary material).}
\captionsetup[subfigure]{labelformat=empty,skip=0pt,position=bottom}
\begin{subfigure}[b]{0.01\linewidth}\phantomsubcaption\label{fig:pca}\end{subfigure}%
\begin{subfigure}[b]{0.01\linewidth}\phantomsubcaption\label{fig:pcadist}\end{subfigure}%
\begin{subfigure}[b]{0.01\linewidth}\phantomsubcaption\label{fig:latency}\end{subfigure}%
\vspace{-14pt}
\end{figure*}

%% file: sec/5_conclusion.tex
\section{Conclusion}
\label{sec:conclusion}

We presented ActMem, an in-context memory framework that decouples memory reasoning from perceptual grounding within a single vision-language backbone.
History-conditioned subtask inference leaves a Latent Memory CoT that encodes what to do, including task progress that the subtask text does not name, while Perception Attention grounds how to act in the current view.
With this design, ActMem reaches 71.1\% task success and 80.8\% completion success on RoboMemArena, 73.8\% on MIKASA-Robo, and 68.8\% on real-world exogenous, endogenous, and traceless memory tasks, above the strongest memory-augmented baselines.

\noindent\textbf{Limitations and Future Work.}
ActMem keeps every sampled frame in its context.
Efficient Memory CoT Inference reuses the history KV cache, but each decision still attends over the full history, so latency grows with episode length, from 0.33\,s to 0.52\,s over the history lengths we test (Sec.~\ref{sec:exp_latency}).
Tasks lasting tens of minutes would instead require selecting which history to keep and retrieving it in latent space, rather than keeping every past frame.

%% file: sec/llm_statement.tex
\section*{LLM Usage Statement}
\label{sec:llm}

We used ChatGPT (OpenAI) to assist with language editing and the organization of the manuscript text, and OpenAI Codex as a coding assistant for parts of the implementation; all such code was reviewed and tested by the authors.
They were not used to generate experimental results or citations.
All technical content, experimental design, data collection, and analyses were carried out by the authors, who verified every claim in this paper and take full responsibility for its content.

%% file: sec/X_suppl.tex
\section{Model and Training Details}
\label{sec:hyper}

Table~\ref{tab:hyper} lists the model and training hyperparameters of ActMem.
ActMem is built on the pretrained starVLA~\cite{starvla} framework.
The vision-language backbone autoregressively generates the subtask for the current decision, and the action expert reads the last-layer hidden states and the token embeddings of the generated subtask tokens as the Latent Memory CoT.
Perception Attention is a zero-initialized, gated cross-attention from the action expert to the current-observation tokens.

\begin{table}[h]
\centering
\caption{Model and training hyperparameters of ActMem.}
\label{tab:hyper}
\resizebox{\linewidth}{!}{%
\begin{tabular}{ll}
\toprule
Hyperparameter & Value \\
\midrule
Vision-language backbone & Qwen3-VL-4B \\
Action expert & Flow-matching DiT, 16 layers \\
Perception Attention layers & 3, 7, 11, 15 \\
Condition for the action expert & Subtask-span hidden states + token embeddings \\
History (RoboMemArena) & 1\,fps, 126\,s \\
History (MIKASA-Robo) & 20\,fps, 3\,s \\
Action chunk / executed actions & 16 / 10 \\
Flow-matching integration steps & 10 \\
Optimizer & AdamW \\
Peak learning rate & $5\times10^{-5}$ (action expert), $5\times10^{-6}$ (VLM backbone) \\
Learning-rate schedule & Cosine decay to $0.1\times$, 1000 warmup steps \\
Global batch size & 64 \\
Training iterations & 60k \\
Precision & bf16 \\
GPUs & 8 $\times$ NVIDIA RTX PRO 6000 \\
\bottomrule
\end{tabular}}
\end{table}

\section{Data and Annotation}
\label{sec:suppl-data}

\paragraph{Subtask annotation.}
For RoboMemArena and MIKASA-Robo, we use the subtask annotations that come with the public LeRobot-format datasets and do not re-annotate.
Each frame carries a subtask index into a per-benchmark list of short English phrases: 69 distinct subtasks over 2,600 demonstrations for RoboMemArena (e.g., ``pick up the chocolate'', ``pour the tomato sauce over the cookies first time''), and 90 over 2,500 demonstrations for MIKASA-Robo (e.g., ``observe the cups'', ``touch the cup hiding the ball'').
For the real-robot tasks, subtask labels are first generated by Qwen3-VL and then spot-checked manually.
For each demonstration, we sample frames at 1\,fps (at most 12 per episode) with their timestamps and prompt Qwen3-VL with the task instruction and a closed vocabulary of phrases.
The model segments the episode into consecutive stages, each named by exactly one phrase from the vocabulary; names outside the vocabulary are discarded.
The vocabularies are:
Cup-and-Block (170 demonstrations), ``find the cup'' and ``pick up the cup'';
Cabinet Search (190), ``open the drawer'', ``pick up the block'', and ``close the drawer'', where ``open the drawer'' repeats once per searched compartment;
Tissue Counting (100), ``approach the tissue box'' and ``pull a tissue first/second/third time'';
Scissor Retrieval (400), ``approach the ottoman'', ``approach the table'', ``move the book aside'', and ``pick up the scissors''.

\paragraph{Prompt template.}
The backbone receives one chat turn.
For RoboMemArena, the user turn contains, in order, the text ``Front camera:'' followed by the timestamped history video (1\,fps, up to 126\,s), the text ``Wrist camera:'' followed by the current image, and the text prompt below.
\begin{quote}\small
\textit{The robot is a Franka Panda single-arm robot. The control frequency is 20 Hz. The next 16 control actions are predicted from the current sub-task. You are given the recent history video from one camera (each frame is tagged with its timestamp in seconds), plus the current image from one more camera. The overall task is: \textless instruction\textgreater. Identify the current sub-task.}
\end{quote}
The assistant turn is the subtask phrase itself (e.g., ``pick up the chocolate''), so the subtask span is the generated answer, and its hidden states form the Latent Memory CoT.
At inference, the same prompt is used and the subtask is decoded greedily.
For MIKASA-Robo, the robot is described as ``a Panda single-arm robot'' and the history is sampled at 20\,fps over 3\,s.

\section{Evaluation Details}
\label{sec:suppl-eval}

\paragraph{RoboMemArena.}
We follow the official protocol with 26 tasks and 51 trials per task, using fixed environment seeds starting from 50.
TSR is the fraction of episodes in which every subtask stage is completed, and CSR is the average fraction of stages completed per episode.
We report the final checkpoint. All ActMem results in Table~1 of the main paper are obtained with Efficient Memory CoT Inference.

\paragraph{Ablation protocol.}
The ablation variants in Sec.~4.4 of the main paper share the data, recipe, and evaluation of the full model and differ only in one switch each:
\textit{w/o History} keeps only the current frame in place of the history video;
\textit{10\,s History} keeps the most recent 10 frames;
\textit{Text-only Subtask} replaces the hidden states of the subtask span with their token embeddings;
\textit{Full-token Conditioning} lets the action expert read all prompt tokens except the answer span;
\textit{w/o Perception Attention} removes the present-observation cross-attention.
All variants are trained for 40k iterations with the learning-rate schedule scaled proportionally, and evaluated on 26 tasks with 10 trials per task.

\section{Real-Robot Setup}
\label{sec:suppl-real}

\begin{figure}[t]
\centering
\begin{minipage}[b]{0.48\linewidth}\centering
\includegraphics[width=0.58\linewidth]{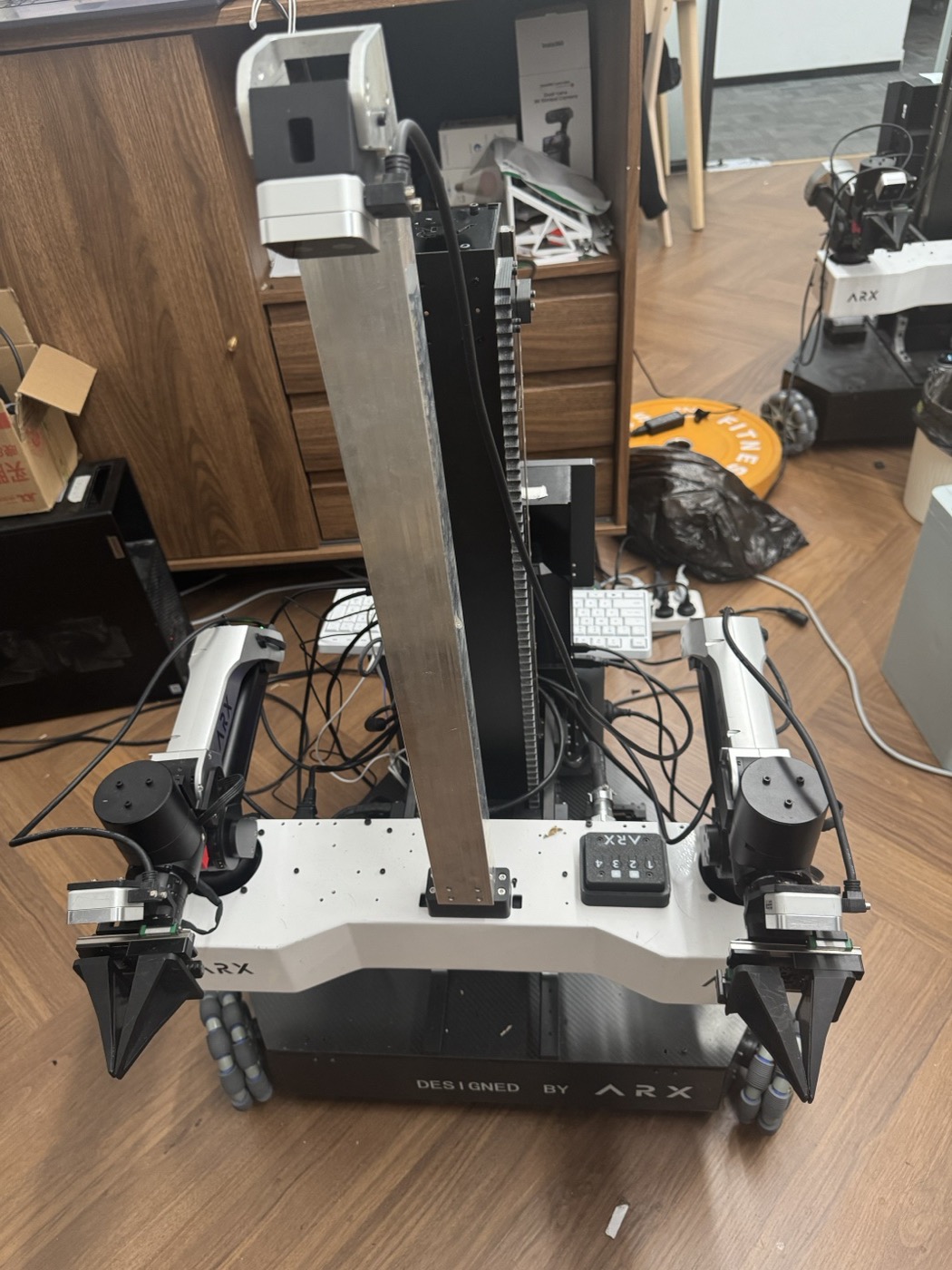}\\[2pt]{\small (a) ARX LIFT2 robot}
\end{minipage}\hfill
\begin{minipage}[b]{0.48\linewidth}\centering
\includegraphics[width=0.58\linewidth]{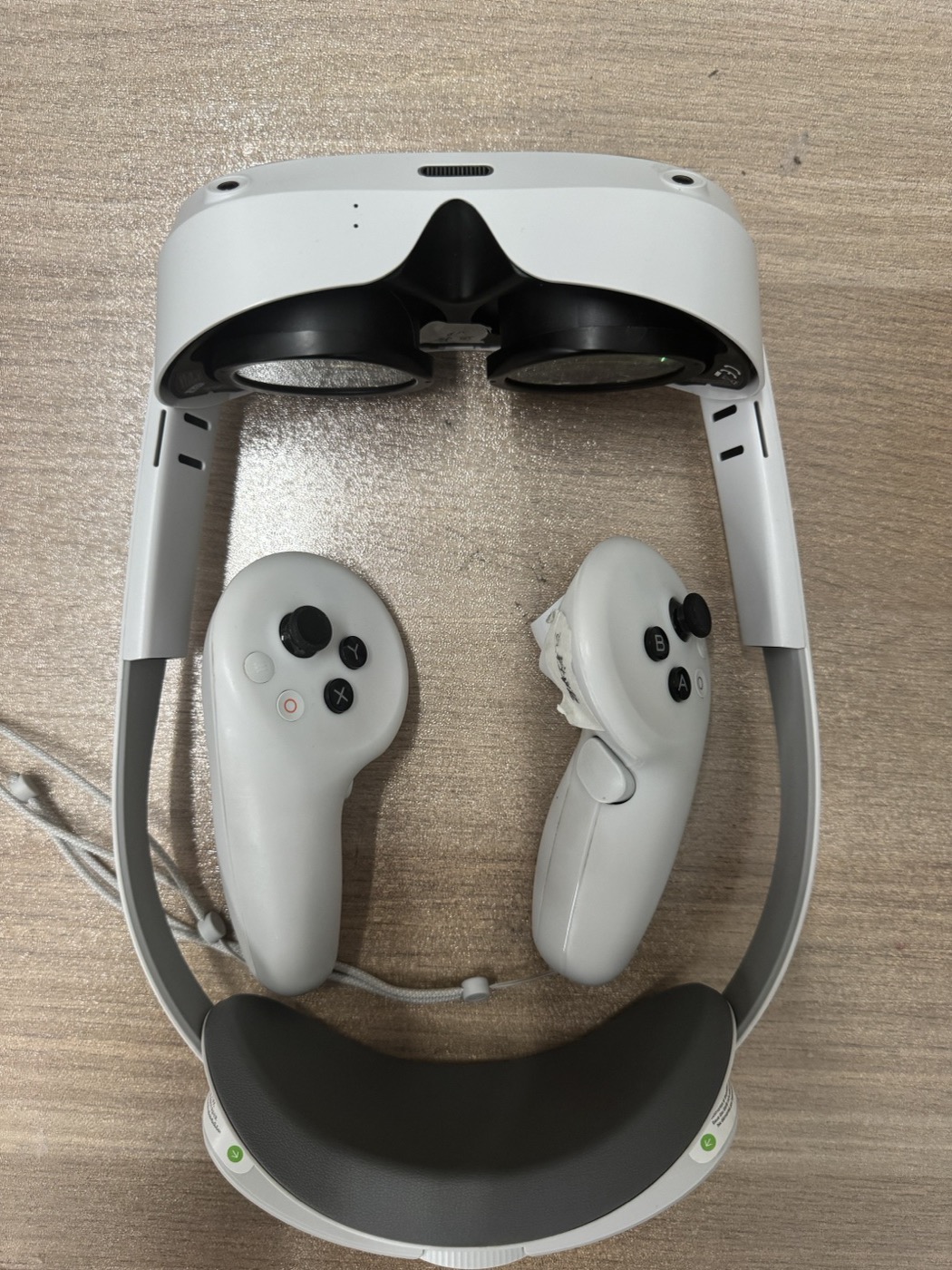}\\[2pt]{\small (b) VR teleoperation device}
\end{minipage}
\caption{Real-robot platform. (a) The ARX LIFT2 dual-arm robot used for all real-world tasks. (b) The VR headset and controllers used to collect demonstrations.}
\label{fig:suppl-platform}
\end{figure}

\paragraph{Platform.}
All real-world experiments use an ARX LIFT2 dual-arm mobile manipulator (Fig.~\ref{fig:suppl-platform}a), which carries two 6-DoF ARX X5 arms with parallel grippers on a lifting column and an omnidirectional wheeled base.
Three Intel RealSense D405 cameras provide the visual input: one head camera mounted on top of the column and one camera on each wrist, all streaming 640$\times$480 RGB at 60 fps.
Demonstrations are collected by VR teleoperation (Fig.~\ref{fig:suppl-platform}b): the operator wears a VR headset and moves two handheld controllers, whose poses are converted to joint commands for the two arms by inverse kinematics and smoothing, and the gripper opening is controlled from the controllers.
At test time, the robot streams its observations to a remote inference station with one NVIDIA RTX 5090 D GPU (24\,GB) over a wired local network, and the policy returns action chunks of 50 steps that are executed at 30\,Hz.

\paragraph{Tasks.}
We evaluate four real-world tasks spanning exogenous, endogenous, and traceless memory.
The language instructions are as follows:
\begin{itemize}
\item \textbf{Cup-and-Block}: ``Retrieve the block after the cups are washed.''
\item \textbf{Cabinet Search}: ``Find the block; do not reopen drawers already inspected.''
\item \textbf{Tissue Counting}: ``Pull tissues the required number of times.''
\item \textbf{Scissor Retrieval}: ``Find the scissors after they have been covered.''
\end{itemize}

\paragraph{Evaluation.}
Each task is evaluated over 40 trials per method, and we report the success rate (1 for success, 0 for failure).

\section{Additional Real-World Visualization}
\label{sec:suppl-vis}

Figs.~\ref{fig:suppl-real-cup}--\ref{fig:suppl-real-scissor} show successful real-robot rollouts of ActMem on the four real-world tasks, one episode per page (two for Tissue Counting), with frames sampled uniformly over each episode.
\input{fig/suppl_realworld}

\section{Inference Latency}
\label{sec:suppl-latency}

\paragraph{Measurement protocol.}
All latencies are measured on one NVIDIA RTX PRO 6000 GPU with bf16 and batch size 1, in the same session.
For ActMem and SimpleMemVLA~\cite{simplememvla}, we run a closed-loop RoboMemArena episode (one task, seed 50, 250 decisions, history clip of 2 to 126 frames at 1\,fps), drop the first five decisions as warm-up, and report the median and 90th percentile of the model time per decision, grouped by the number of history frames.
ActMem runs both inference modes back to back on the same observation with the same random state, and the rollout follows the actions of standard inference.
Standard inference re-encodes the full history, whereas Efficient Memory CoT Inference reuses cached vision features and the longest matching key-value prefix and re-decodes the previous subtask only when it no longer matches.
Times are wall-clock with CUDA synchronization and exclude building observation features.
For SimpleMemVLA, the reported time covers the model call without the input processor, namely the image encoder, subtask generation, and the 10-step action head.
The $\pi_{0.5}$-family models (the $\pi_{0.5}$ base, MEM-$\pi_{0.5}$ with short and long memory, and our reimplementation of MemoryVLA on $\pi_{0.5}$) keep a fixed-size memory, so we report one value per model, measured offline in JAX on the same GPU over 50 calls with 10 flow-matching steps, three $224\times224$ camera inputs, and memory filled to capacity.

\begin{table}[h]
\centering
\caption{Median / 90th-percentile per-decision latency (s) by history length, measured on the same GPU in the same session.}
\label{tab:latency}
\small
\setlength{\tabcolsep}{4pt}
\resizebox{\linewidth}{!}{%
\begin{tabular}{lcccc}
\toprule
History (frames) & 0--30 & 30--60 & 60--90 & 90--126 \\
Decisions & 51 & 60 & 60 & 74 \\
\midrule
SimpleMemVLA & 0.70 / 1.05 & 0.85 / 1.29 & 1.02 / 1.67 & 1.25 / 1.96 \\
ActMem, standard & 0.87 / 1.45 & 0.98 / 1.68 & 1.09 / 1.94 & 1.42 / 2.27 \\
\rowcolor{evkBlue!45}ActMem, Efficient Memory CoT & 0.33 / 0.52 & 0.37 / 0.56 & 0.52 / 0.75 & 0.50 / 0.74 \\
\bottomrule
\end{tabular}}
\end{table}

\begin{table}[h]
\centering
\caption{Median / 90th-percentile latency (ms) of $\pi_{0.5}$-family models with a fixed-size memory, measured offline on the same GPU.}
\label{tab:latency_pi}
\small
\setlength{\tabcolsep}{6pt}
\resizebox{\linewidth}{!}{%
\begin{tabular}{lc}
\toprule
Model & Latency (ms) \\
\midrule
$\pi_{0.5}$~\cite{pi05} & 78.6 / 88.8 \\
MEM-$\pi_{0.5}$ (short memory)~\cite{mem} & 93.6 / 123.1 \\
MEM-$\pi_{0.5}$ (long memory)~\cite{mem} & 78.3 / 122.2 \\
MemoryVLA~\cite{memoryvla} (reimplemented on $\pi_{0.5}$) & 94.8 / 103.3 \\
\bottomrule
\end{tabular}}
\end{table}

\paragraph{Where the time goes.}
With Efficient Memory CoT Inference, the key-value prefix is reused in 98.4\% of decisions, exactly one new vision chunk is encoded per decision, and the previous subtask is accepted without decoding in 71.2\% of decisions (94\% up to 60 frames of history, 51\% beyond).
The action expert takes about 0.16\,s regardless of history length.

%% file: fig/suppl_realworld.tex
\newcommand{\realpage}[3]{%
\begin{figure*}[p]
\centering
\includegraphics[width=\textwidth,height=0.9\textheight,keepaspectratio]{fig/suppl_real/#1.jpg}
\caption{#2}
\label{#3}
\end{figure*}}

\realpage{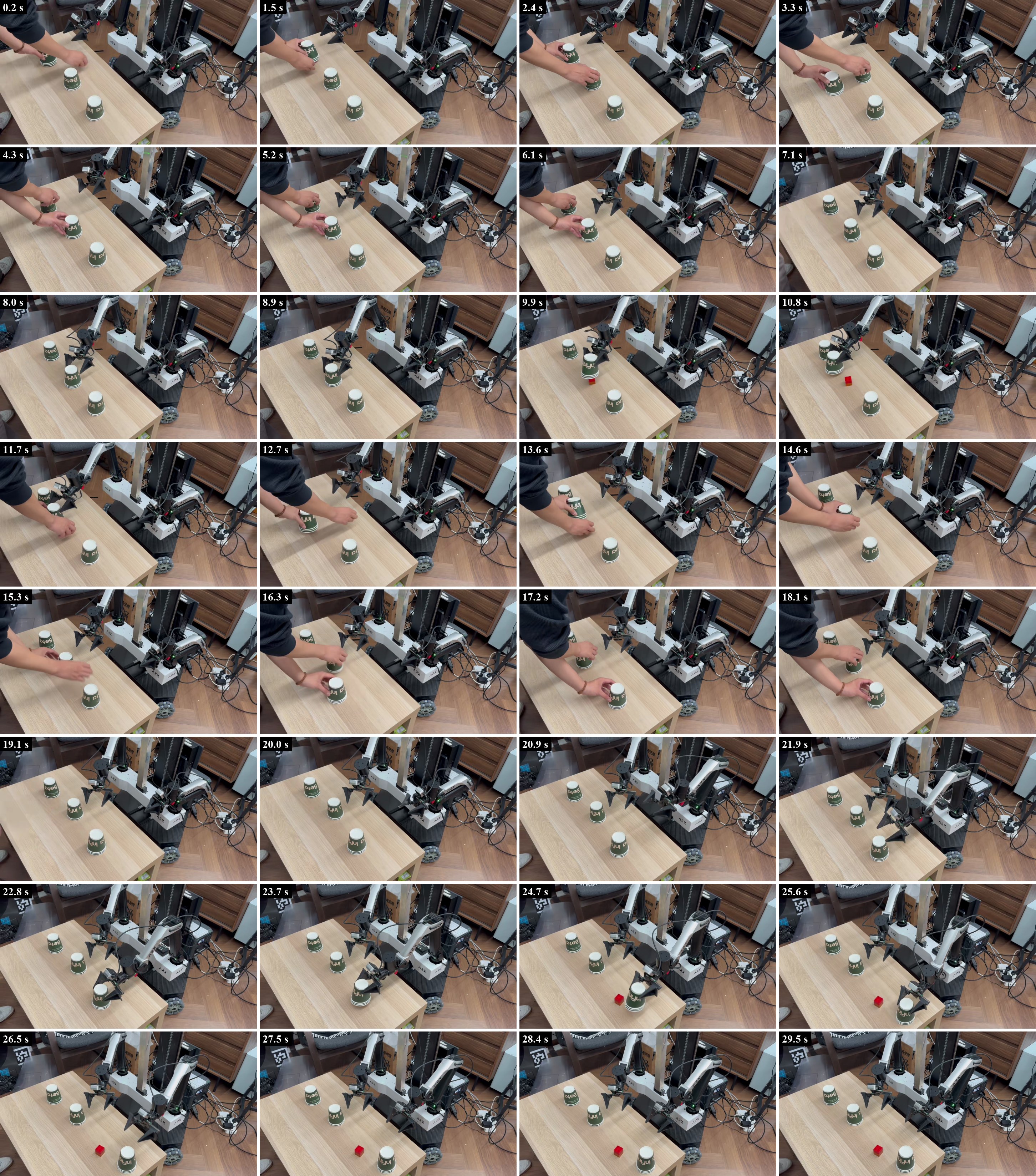}{Successful real-robot rollout of ActMem on \textbf{Cup-and-Block} (``Retrieve the block after the cups are washed.''). Frames are sampled uniformly over the 30\,s episode in raster order; the timestamp is shown at the top-left of each frame.}{fig:suppl-real-cup}

\realpage{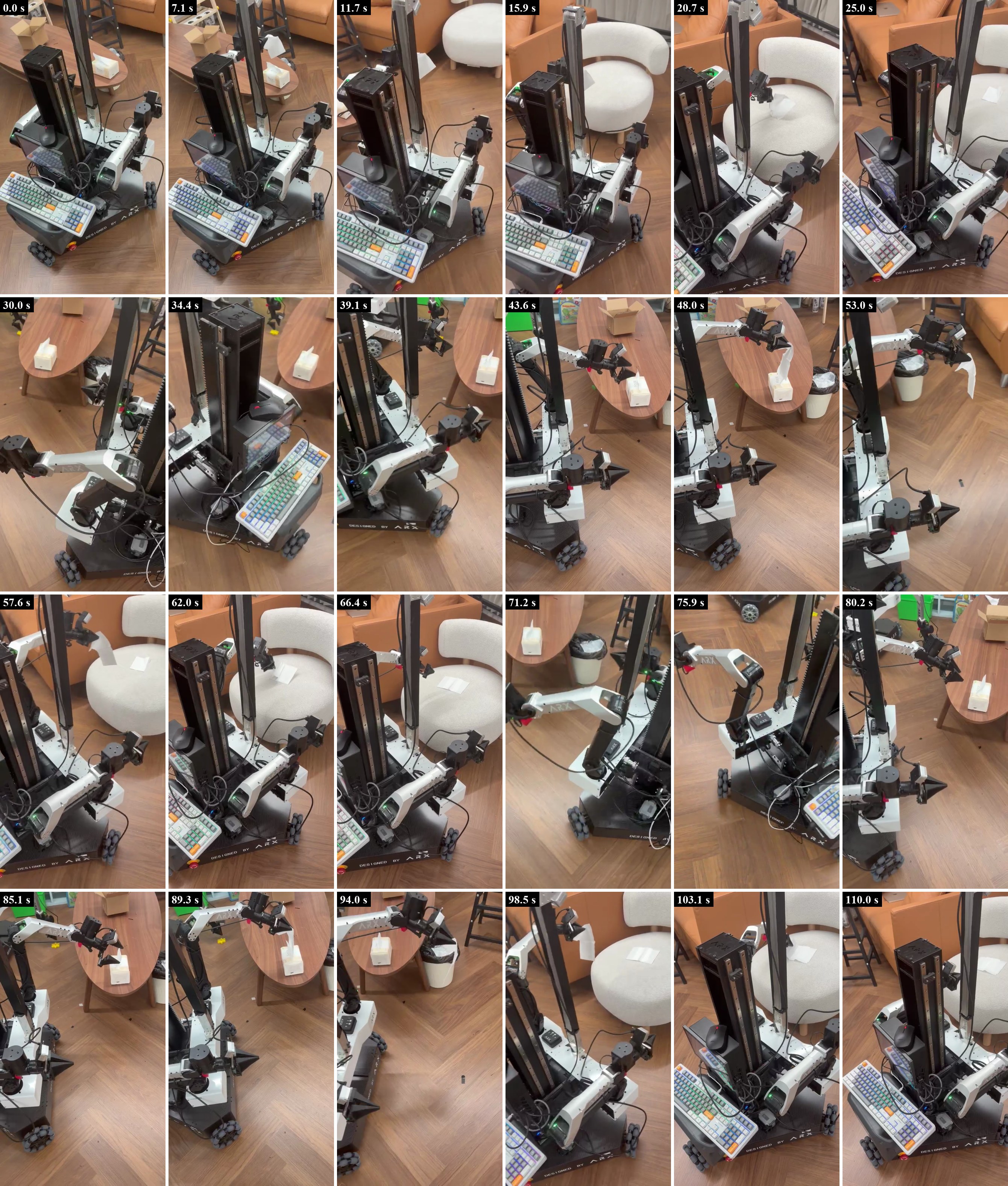}{Successful real-robot rollout of ActMem on \textbf{Tissue Counting} (``Pull tissues the required number of times.''), episode 1. Frames are sampled uniformly over the 110\,s episode in raster order.}{fig:suppl-real-tissue1}

\realpage{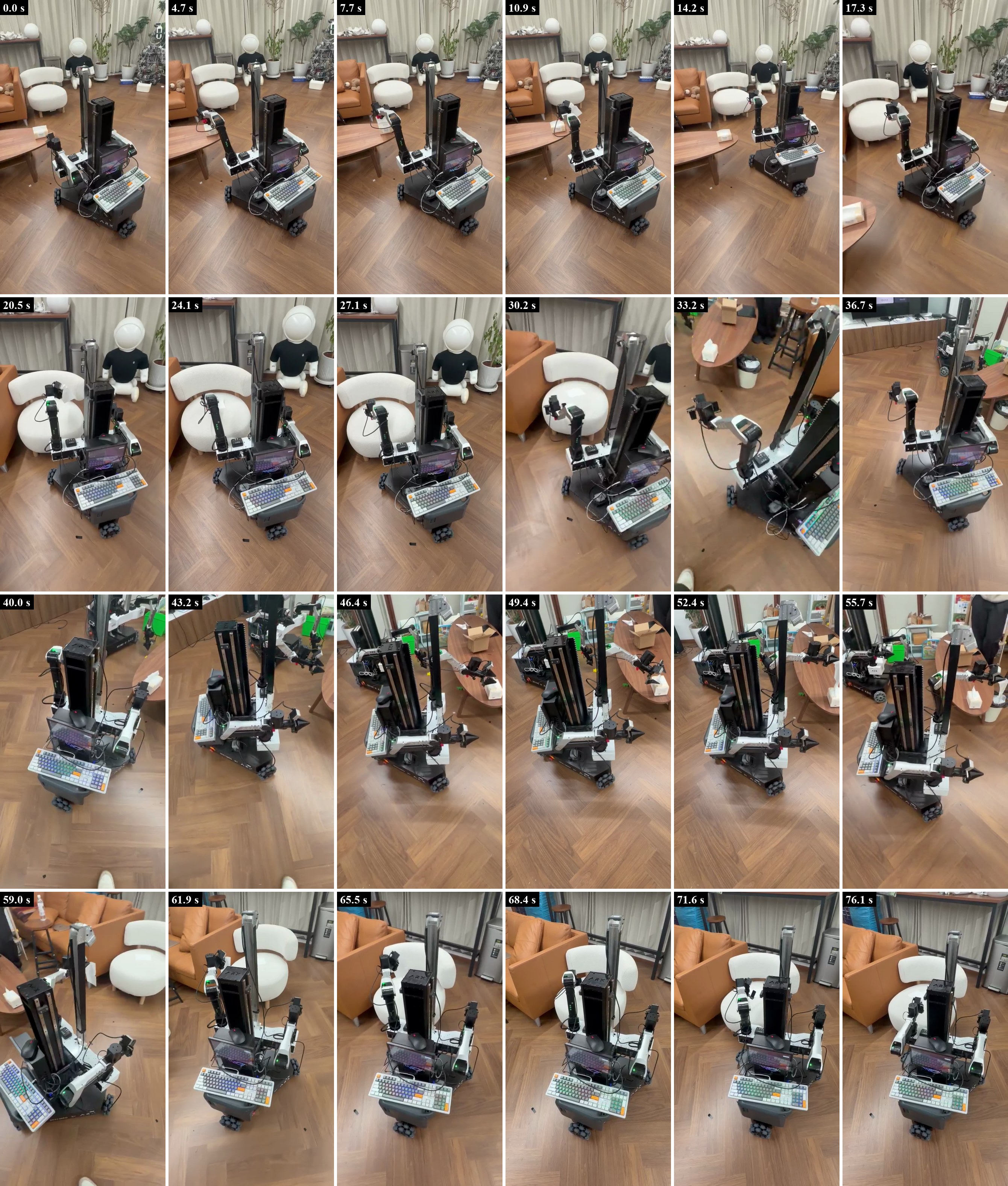}{Successful real-robot rollout of ActMem on \textbf{Tissue Counting}, episode 2. Frames are sampled uniformly over the 77\,s episode in raster order.}{fig:suppl-real-tissue2}

\realpage{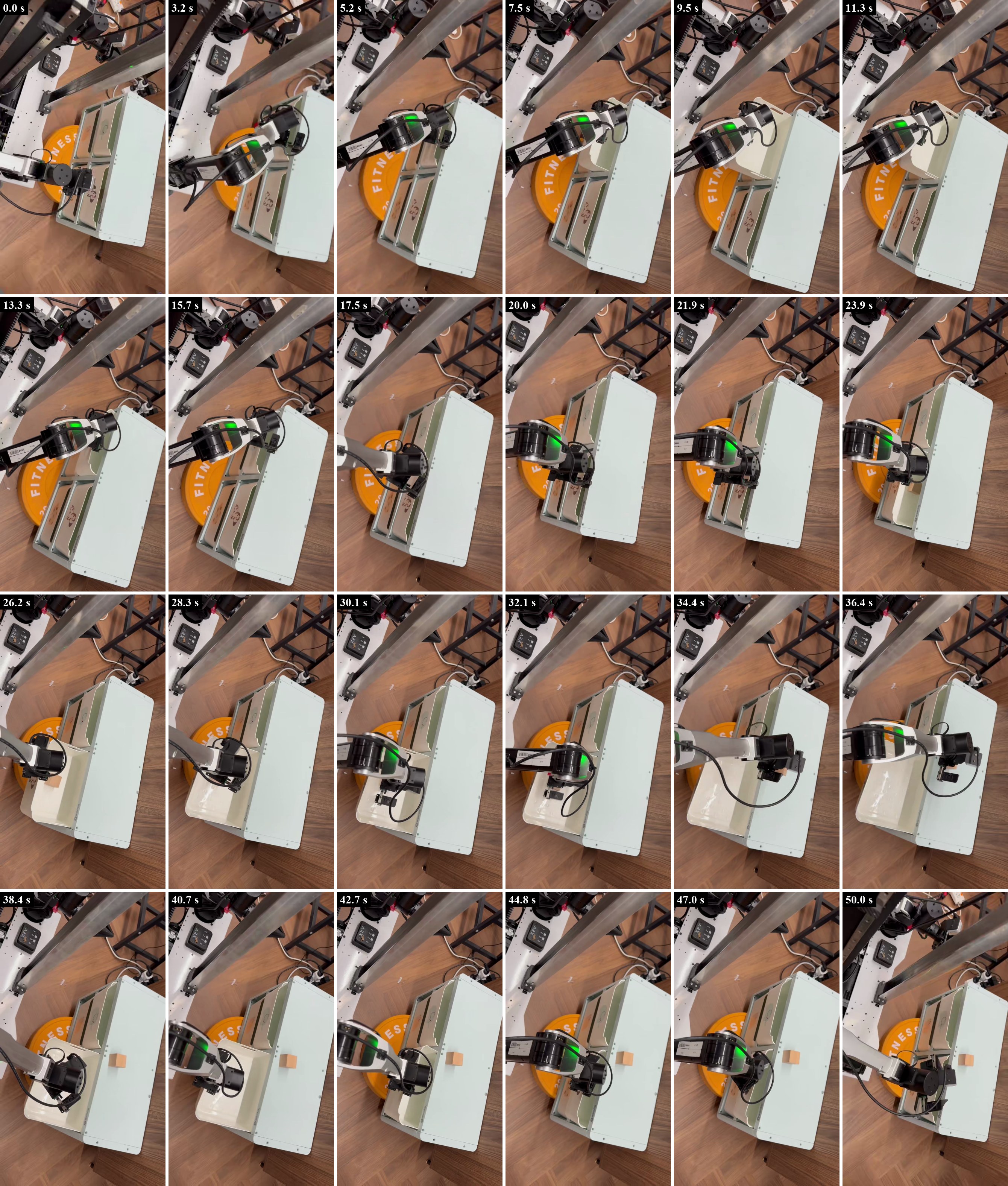}{Successful real-robot rollout of ActMem on \textbf{Cabinet Search} (``Find the block; do not reopen drawers already inspected.''). Frames are sampled uniformly over the 50\,s episode in raster order.}{fig:suppl-real-cabinet}

\realpage{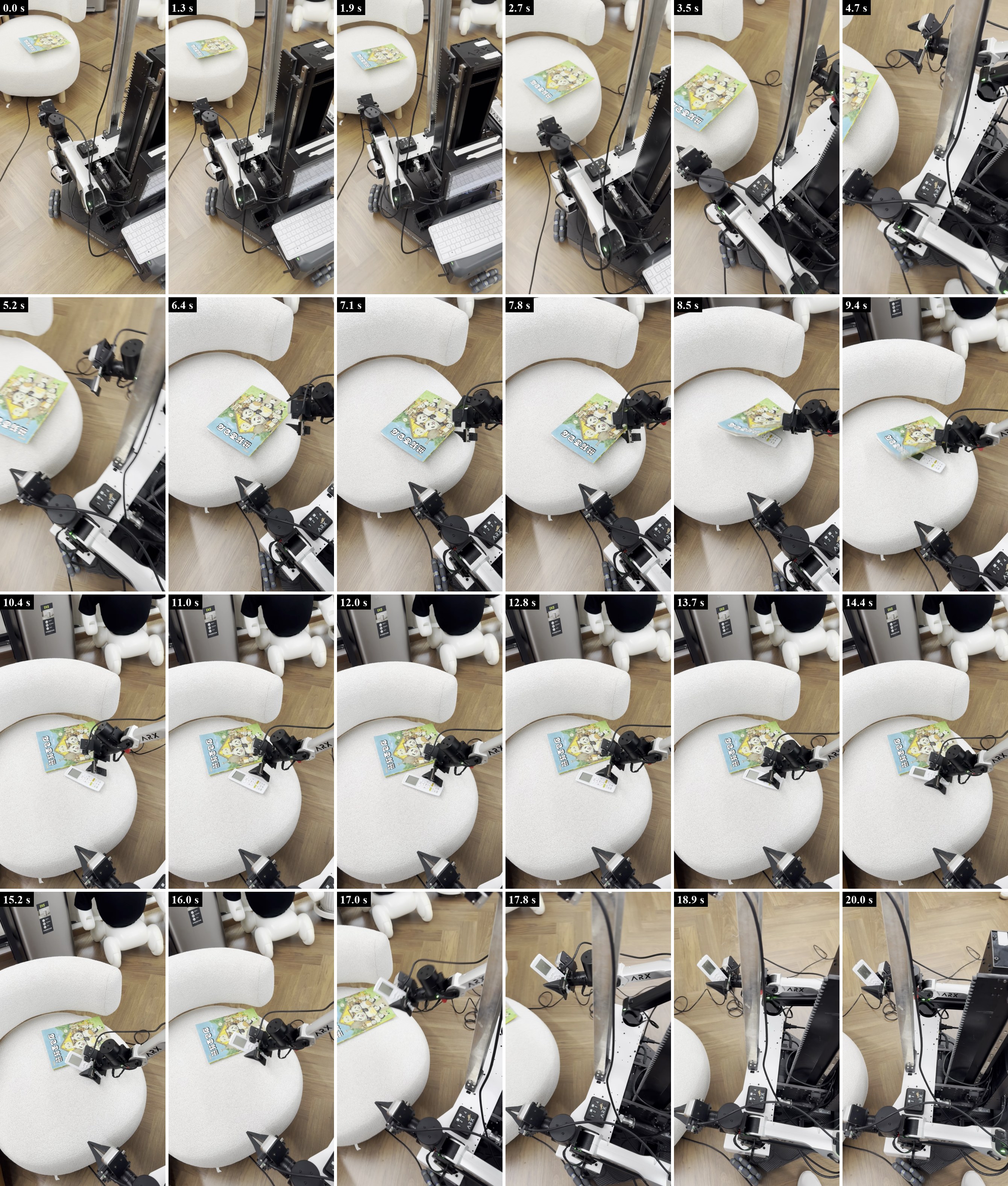}{Successful real-robot rollout of ActMem on \textbf{Scissor Retrieval} (``Find the scissors after they have been covered.''). Frames are sampled uniformly over the 20\,s episode in raster order.}{fig:suppl-real-scissor}

%% file: full.bbl
\begin{thebibliography}{40}
\providecommand{\natexlab}[1]{#1}
\providecommand{\url}[1]{\texttt{#1}}
\expandafter\ifx\csname urlstyle\endcsname\relax
  \providecommand{\doi}[1]{doi: #1}\else
  \providecommand{\doi}{doi: \begingroup \urlstyle{rm}\Url}\fi

\bibitem[Bai et~al.(2025{\natexlab{a}})Bai, Cai, Chen, Chen, Chen, Cheng, Deng, Ding, Fang, Gao, et~al.]{qwen3vl}
Shuai Bai, Yuxuan Cai, Rui-Zhe Chen, Ke-qin Chen, Xiong-Hui Chen, Zesen Cheng, Lianghao Deng, Wei Ding, Rongyao Fang, Chang Gao, et~al.
\newblock {Qwen3-VL} technical report.
\newblock \emph{arXiv.org}, 2025{\natexlab{a}}.

\bibitem[Bai et~al.(2025{\natexlab{b}})Bai, Chen, Liu, Wang, Ge, Song, Dang, Wang, Wang, Tang, et~al.]{qwen25vl}
Shuai Bai, Ke-qin Chen, Xuejing Liu, Jialin Wang, Wenbin Ge, Sibo Song, K. Dang, Peng Wang, Shijie Wang, Jun Tang, et~al.
\newblock {Qwen2.5-VL} technical report.
\newblock \emph{arXiv.org}, 2025{\natexlab{b}}.

\bibitem[Beyer et~al.(2024)Beyer, Steiner, Pinto, Kolesnikov, Wang, Salz, Neumann, Alabdulmohsin, Tschannen, Bugliarello, et~al.]{paligemma}
Lucas Beyer, Andreas Steiner, Andr{\'e}~Susano Pinto, Alexander Kolesnikov, Xiao Wang, Daniel Salz, Maxim Neumann, Ibrahim Alabdulmohsin, Michael Tschannen, Emanuele Bugliarello, et~al.
\newblock Paligemma: A versatile 3b vlm for transfer.
\newblock \emph{arXiv preprint arXiv:2407.07726}, 2024.

\bibitem[Bjorck et~al.(2025)Bjorck, Castaneda, Cherniadev, et~al.]{gr00t}
Johan Bjorck, Fernando Castaneda, Nikita Cherniadev, et~al.
\newblock {GR00T N1}: An open foundation model for generalist humanoid robots.
\newblock \emph{arXiv preprint arXiv:2503.14734}, 2025.

\bibitem[Black et~al.(2024)Black, Brown, Driess, Esmail, Equi, Finn, Fusai, Groom, Hausman, Ichter, et~al.]{pi0}
Kevin Black, Noah Brown, Danny Driess, Adnan Esmail, Michael Equi, Chelsea Finn, Niccolo Fusai, Lachy Groom, Karol Hausman, Brian Ichter, et~al.
\newblock {$\pi_0$: A Vision-Language-Action Flow Model for General Robot Control}, 2024.

\bibitem[Chen et~al.(2026)Chen, Wang, Li, Qin, Shi, Li, Hu, Zhang, Wang, Chen, Wang, Xu, Wu, Mu, Yang, Dong, and Luo]{rmbench}
Tianxing Chen, Yuran Wang, Mingleyang Li, Yan Qin, Hao Shi, Zixuan Li, Yifan Hu, Yingsheng Zhang, Kaixuan Wang, Yue Chen, Hongcheng Wang, Renjing Xu, Ruihai Wu, Yao Mu, Yaodong Yang, Hao Dong, and Ping Luo.
\newblock Rmbench: Memory-dependent robotic manipulation benchmark with insights into policy design, 2026.

\bibitem[Cherepanov et~al.(2025)Cherepanov, Kachaev, Kovalev, and Panov]{mikasa}
E. Cherepanov, Nikita Kachaev, A. Kovalev, and Aleksandr~I. Panov.
\newblock Memory, benchmark \& robots: A benchmark for solving complex tasks with reinforcement learning.
\newblock \emph{arXiv preprint arXiv:2502.10550}, 2025.

\bibitem[Community(2026)]{starvla}
StarVLA Community.
\newblock Starvla: A lego-like codebase for vision-language-action model developing.
\newblock \emph{arXiv preprint arXiv:2604.05014}, 2026.

\bibitem[Dai et~al.(2026)Dai, Fu, Lee, Liu, Zhang, Yang, Finn, Fazeli, and Chai]{robomme}
Yinpei Dai, Hongze Fu, Jayjun Lee, Yuejiang Liu, Haoran Zhang, Jianing Yang, Chelsea Finn, Nima Fazeli, and Joyce Chai.
\newblock Robomme: Benchmarking and understanding memory for robotic generalist policies, 2026.

\bibitem[Fang et~al.(2025)Fang, Grotz, Pumacay, Wang, Fox, Krishna, and Duan]{sam2act}
Haoquan Fang, Markus Grotz, Wilbert Pumacay, Yi~Ru Wang, Dieter Fox, Ranjay Krishna, and Jiafei Duan.
\newblock Sam2act: Integrating visual foundation model with a memory architecture for robotic manipulation.
\newblock \emph{arXiv preprint arXiv:2501.18564}, 2025.

\bibitem[Gao et~al.(2026)Gao, Liu, Li, and Song]{gmp}
Yihuai Gao, Jeff~Jinyun Liu, Shuang Li, and Shuran Song.
\newblock Gated memory policy: In-context memorization and adaptation.
\newblock \emph{arXiv preprint arXiv:2604.18933}, 2026.

\bibitem[Goodale and Milner(1992)]{goodale1992}
Melvyn~A. Goodale and A.~David Milner.
\newblock Separate visual pathways for perception and action.
\newblock \emph{Trends in Neurosciences}, 15\penalty0 (1):\penalty0 20--25, 1992.

\bibitem[Intelligence et~al.(2025)Intelligence, Black, Brown, Darpinian, Dhabalia, Driess, Esmail, Equi, Finn, Fusai, et~al.]{pi05}
Physical Intelligence, Kevin Black, Noah Brown, James Darpinian, Karan Dhabalia, Danny Driess, Adnan Esmail, Michael Equi, Chelsea Finn, Niccolo Fusai, et~al.
\newblock $\pi_{0.5}$: a vision-language-action model with open-world generalization.
\newblock \emph{arXiv preprint arXiv:2504.16054}, 2025.

\bibitem[Jang et~al.(2025)Jang, Yu, Kwon, Jeon, Seo, and Shin]{contextvla}
Huiwon Jang, Sihyun Yu, Heeseung Kwon, Hojin Jeon, Younggyo Seo, and Jinwoo Shin.
\newblock Contextvla: Vision-language-action model with amortized multi-frame context.
\newblock \emph{arXiv preprint arXiv:2510.04246}, 2025.

\bibitem[Kaelbling et~al.(1998)Kaelbling, Littman, and Cassandra]{kaelbling1998pomdp}
Leslie~Pack Kaelbling, Michael~L. Littman, and Anthony~R. Cassandra.
\newblock Planning and acting in partially observable stochastic domains.
\newblock \emph{Artificial Intelligence}, 101\penalty0 (1--2):\penalty0 99--134, 1998.

\bibitem[Kim et~al.(2024)Kim, Pertsch, Karamcheti, Xiao, Balakrishna, Nair, Rafailov, Foster, Lam, Sanketi, Vuong, Kollar, Burchfiel, Tedrake, Sadigh, Levine, Liang, and Finn]{openvla}
Moo~Jin Kim, Karl Pertsch, Siddharth Karamcheti, Ted Xiao, Ashwin Balakrishna, Suraj Nair, Rafael Rafailov, Ethan Foster, Grace Lam, Pannag Sanketi, Quan Vuong, Thomas Kollar, Benjamin Burchfiel, Russ Tedrake, Dorsa Sadigh, Sergey Levine, Percy Liang, and Chelsea Finn.
\newblock Openvla: An open-source vision-language-action model, 2024.

\bibitem[Kim et~al.(2025)Kim, Finn, and Liang]{openvlaoft}
Moo~Jin Kim, Chelsea Finn, and Percy Liang.
\newblock Fine-tuning vision-language-action models: Optimizing speed and success.
\newblock \emph{RSS}, 2025.

\bibitem[Koo et~al.(2025)Koo, Choi, Kim, Lee, Kim, Seo, and Shin]{hamlet}
Myungkyu Koo, Daewon Choi, Taeyoung Kim, Kyungmin Lee, Changyeon Kim, Younggyo Seo, and Jinwoo Shin.
\newblock Hamlet: Switch your vision-language-action model into a history-aware policy.
\newblock \emph{arXiv preprint arXiv:2510.00695}, 2025.

\bibitem[Lei et~al.(2026)Lei, Song, Zhang, Pei, Chen, Yan, Zhao, Ding, Zhang, Huang, Wang, Wang, and Li]{robomemarena}
Huashuo Lei, Wenxuan Song, Huarui Zhang, Jieyuan Pei, Jiayi Chen, Haodong Yan, Han Zhao, Pengxiang Ding, Zhipeng Zhang, Lida Huang, Donglin Wang, Yan Wang, and Haoang Li.
\newblock Robomemarena: A comprehensive and challenging robotic memory benchmark, 2026.

\bibitem[Leviathan et~al.(2023)Leviathan, Kalman, and Matias]{leviathan2023speculative}
Yaniv Leviathan, Matan Kalman, and Yossi Matias.
\newblock Fast inference from transformers via speculative decoding.
\newblock In \emph{International Conference on Machine Learning}, 2023.

\bibitem[Li et~al.(2025)Li, Yang, Chen, Tian, Yang, Chen, Wang, Wang, Zhao, Lin, et~al.]{cronusvla}
Hao Li, Shuai Yang, Yilun Chen, Yang Tian, Xiaoda Yang, Xinyi Chen, Hanqing Wang, Tai Wang, Feng Zhao, Dahua Lin, et~al.
\newblock {CronusVLA}: Towards efficient and robust manipulation via multi-frame vision-language-action modeling.
\newblock \emph{arXiv preprint arXiv:2506.19816}, 2025.

\bibitem[Li et~al.(2024)Li, Liang, Wang, Luo, Chen, Liao, Wei, Deng, Xu, Zhang, et~al.]{cogact}
Qixiu Li, Yaobo Liang, Zeyu Wang, Lin Luo, Xi Chen, Mozheng Liao, Fangyun Wei, Yu Deng, Sicheng Xu, Yizhong Zhang, et~al.
\newblock Cogact: A foundational vision-language-action model for synergizing cognition and action in robotic manipulation.
\newblock \emph{arXiv preprint arXiv:2411.19650}, 2024.

\bibitem[Lin et~al.(2025)Lin, Ding, Wang, Zhuang, Liu, Tong, Song, Lyu, Huang, and Wang]{hifvla}
Minghui Lin, Pengxiang Ding, Shu Wang, Zifeng Zhuang, Yang Liu, Xinyang Tong, Wenxuan Song, Shangke Lyu, Siteng Huang, and Donglin Wang.
\newblock Hif-vla: Hindsight, insight and foresight through motion representation for vision-language-action models.
\newblock \emph{arXiv preprint arXiv:2512.09928}, 2025.

\bibitem[Liu et~al.(2026)Liu, Huang, Hong, Du, Cao, Fei-Fei, and Wu]{t2mem}
Yize Liu, Huang Huang, Yining Hong, Zijian Du, Zhi Cao, Li Fei-Fei, and Jiajun Wu.
\newblock {T}$^2${Mem}: Learning test-time memory for robotics.
\newblock \emph{arXiv preprint arXiv:2609.36720}, 2026.

\bibitem[Qu et~al.(2025)Qu, Song, Chen, Yao, Ye, Ding, Wang, Gu, Zhao, Wang, et~al.]{spatialvla}
Delin Qu, Haoming Song, Qizhi Chen, Yuanqi Yao, Xinyi Ye, Yan Ding, Zhigang Wang, JiaYuan Gu, Bin Zhao, Dong Wang, et~al.
\newblock {SpatialVLA}: Exploring spatial representations for visual-language-action models.
\newblock In \emph{Robotics: Science and Systems}, 2025.

\bibitem[Schacter et~al.(2007)Schacter, Addis, and Buckner]{schacter2007}
Daniel~L. Schacter, Donna~Rose Addis, and Randy~L. Buckner.
\newblock Remembering the past to imagine the future: the prospective brain.
\newblock \emph{Nature Reviews Neuroscience}, 8\penalty0 (9):\penalty0 657--661, 2007.

\bibitem[Shi et~al.(2025{\natexlab{a}})Shi, Xie, Liu, Sun, Liu, Wang, Zhou, Fan, Zhang, and Huang]{memoryvla}
Hao Shi, Bin Xie, Yingfei Liu, Lin Sun, Fengrong Liu, Tiancai Wang, Erjin Zhou, Haoqiang Fan, Xiangyu Zhang, and Gao Huang.
\newblock Memoryvla: Perceptual-cognitive memory in vision-language-action models for robotic manipulation.
\newblock \emph{arXiv preprint arXiv:2508.19236}, 2025{\natexlab{a}}.

\bibitem[Shi et~al.(2026)Shi, Li, Xie, Wang, Zhou, Wang, Zhang, Luo, and Huang]{memoryvlapp}
Hao Shi, Weiye Li, Bin Xie, Yulin Wang, Renping Zhou, Tiancai Wang, Xiangyu Zhang, Ping Luo, and Gao Huang.
\newblock Memoryvla++: Temporal modeling via memory and imagination in vision-language-action models, 2026.

\bibitem[Shi et~al.(2025{\natexlab{b}})Shi, Ichter, Equi, Ke, Pertsch, Vuong, Tanner, Walling, Wang, Fusai, et~al.]{hirobot}
Lucy~Xiaoyang Shi, Brian Ichter, Michael Equi, Liyiming Ke, Karl Pertsch, Quan Vuong, James Tanner, Anna Walling, Haohuan Wang, Niccolo Fusai, et~al.
\newblock {Hi Robot}: Open-ended instruction following with hierarchical vision-language-action models.
\newblock \emph{arXiv preprint arXiv:2502.19417}, 2025{\natexlab{b}}.

\bibitem[Sridhar et~al.(2025)Sridhar, Pan, Sharma, and Finn]{memer}
Ajay Sridhar, Jennifer Pan, Satvik Sharma, and Chelsea Finn.
\newblock Memer: Scaling up memory for robot control via experience retrieval.
\newblock \emph{arXiv preprint arXiv:2510.20328}, 2025.

\bibitem[Team et~al.(2024)Team, Ghosh, Walke, Pertsch, Black, Mees, Dasari, Hejna, Kreiman, Xu, et~al.]{octo}
Octo~Model Team, Dibya Ghosh, Homer Walke, Karl Pertsch, Kevin Black, Oier Mees, Sudeep Dasari, Joey Hejna, Tobias Kreiman, Charles Xu, et~al.
\newblock Octo: An open-source generalist robot policy.
\newblock In \emph{Robotics: Science and Systems}, 2024.

\bibitem[Torne et~al.(2026)Torne, Pertsch, Walke, Vedder, Nair, Ichter, Ren, Wang, Tang, Stachowicz, Dhabalia, Equi, Vuong, Springenberg, Levine, Finn, and Driess]{mem}
Marcel Torne, Karl Pertsch, Homer Walke, Kyle Vedder, Suraj Nair, Brian Ichter, Allen~Z. Ren, Haohuan Wang, Jiaming Tang, Kyle Stachowicz, Karan Dhabalia, Michael Equi, Quan Vuong, Jost~Tobias Springenberg, Sergey Levine, Chelsea Finn, and Danny Driess.
\newblock {MEM}: Multi-scale embodied memory for vision language action models.
\newblock arXiv preprint arXiv:2603.03596, 2026.

\bibitem[Wang et~al.(2026{\natexlab{a}})Wang, Gu, Chen, Xu, Ma, Su, Li, Huang, and Wang]{dimwam}
Kai Wang, Zhaopeng Gu, Yixiang Chen, Yuan Xu, Qisen Ma, Peng Su, Zhaowen Li, Yan Huang, and Liang Wang.
\newblock Dim-wam: World-action modeling with diverse historical event memory, 2026{\natexlab{a}}.

\bibitem[Wang et~al.(2026{\natexlab{b}})Wang, Shi, Ni, Yang, Li, Su, Lin, and Li]{nativemem}
Ziye Wang, Modi Shi, Chaojun Ni, Jiazhi Yang, Mengdi Li, Zhizhong Su, Tianwei Lin, and Hongyang Li.
\newblock {NativeMEM}: Native memory compression for long-horizon robotic manipulation.
\newblock \emph{arXiv preprint arXiv:2607.06678}, 2026{\natexlab{b}}.

\bibitem[Yang et~al.(2026{\natexlab{a}})Yang, Tu, Yang, Mao, Dong, Chen, Peng, Xiong, Cao, Dai, Zhou, Mu, and Wang]{eventvla}
Ganlin Yang, Zhangzheng Tu, Yuqiang Yang, Sitong Mao, Junyi Dong, Tianxing Chen, Jiaqi Peng, Jing Xiong, Jiafei Cao, Jifeng Dai, Wengang Zhou, Yao Mu, and Tai Wang.
\newblock Eventvla: Event-driven visual evidence memory for long-horizon vision-language-action policies, 2026{\natexlab{a}}.

\bibitem[Yang et~al.(2026{\natexlab{b}})Yang, Mu, Wei, Lu, Li, Xu, Xue, Yuan, Lin, Pang, and Xu]{memorywam}
Sizhe Yang, Juncheng Mu, Tianming Wei, Chenhao Lu, Xiaofan Li, Linning Xu, Zhengrong Xue, Zhecheng Yuan, Dahua Lin, Jiangmiao Pang, and Huazhe Xu.
\newblock Memorywam: Efficient world action modeling with persistent memory, 2026{\natexlab{b}}.

\bibitem[Yin et~al.(2026)Yin, Xu, Yang, Tam, Liu, Yao, Zeng, Cui, Wang, Yin, and Lin]{simplememvla}
Cheng Yin, Wang Xu, Junpeng Yang, Sikyuen Tam, Hanyu Liu, Yuan Yao, Xiangrui Zeng, Junbo Cui, Yequan Wang, Zhouping Yin, and Yankai Lin.
\newblock {SimpleMemVLA}: A simple but effective native-video memory for vision-language-action models.
\newblock \emph{arXiv preprint arXiv:2609.05533}, 2026.

\bibitem[Zeng et~al.(2026)Zeng, Ye, Chen, Shentu, Wu, Yan, and Li]{kemo}
Yihan Zeng, Minghao Ye, Yiyuan Chen, Yide Shentu, Philipp Wu, Zike Yan, and Zhongyu Li.
\newblock {KEMO}: Event-driven keyframe memory for long-horizon robot manipulation with {VLA} policies.
\newblock \emph{arXiv preprint arXiv:2606.23589}, 2026.

\bibitem[Zhong et~al.(2026)Zhong, Liu, Wei, Xiong, Yao, Liu, and Ren]{acotvla}
Linqing Zhong, Yi Liu, Yifei Wei, Ziyu Xiong, Maoqing Yao, Si Liu, and Guanghui Ren.
\newblock {ACoT-VLA}: Action chain-of-thought for vision-language-action models.
\newblock \emph{arXiv preprint arXiv:2601.11404}, 2026.

\bibitem[Zitkovich et~al.(2023)Zitkovich, Yu, Xu, Xu, Xiao, Xia, Wu, Wohlhart, Welker, Wahid, et~al.]{rt2}
Brianna Zitkovich, Tianhe Yu, Sichun Xu, Peng Xu, Ted Xiao, Fei Xia, Jialin Wu, Paul Wohlhart, Stefan Welker, Ayzaan Wahid, et~al.
\newblock Rt-2: Vision-language-action models transfer web knowledge to robotic control.
\newblock In \emph{Conference on Robot Learning}, pages 2165--2183. PMLR, 2023.

\end{thebibliography}
